\documentclass[journal]{IEEEtran}
\usepackage{cite}
\usepackage{amsmath,amssymb,amsfonts}
\usepackage{bm}
\usepackage{algorithmic}
\usepackage{graphicx}
\usepackage{textcomp}
\usepackage{xcolor}
\usepackage[colorlinks=true, linkcolor=black, citecolor=black, urlcolor=black]{hyperref}
\usepackage{bbding}
\usepackage{multirow}
\usepackage{eso-pic}

\usepackage{booktabs}
\usepackage[acronym, shortcuts]{glossaries}
\newacronym{gai}{GenAI}{generative AI}
\newacronym{dl}{DL}{Deep Learning}
\newacronym{roc}{ROC}{Receiver Operating Characteristic}
\newacronym{ba}{BA}{Balanced Accuracy}
\newacronym{nn}{NN}{Neural Network}
\newacronym{cnn}{CNN}{Convolutional Neural Network}
\newacronym{mse}{MSE}{Mean Squared Error}
\newacronym{tp}{TP}{True Positive}
\newacronym{tn}{TN}{True Negative}
\newacronym{fp}{FP}{False Positive}
\newacronym{fn}{FN}{False Negative}
\newacronym{relu}{ReLU}{Rectified Linear Unit}
\newacronym{ml}{ML}{Machine Learning}
\newacronym{rms}{RMS}{Root Mean Square}
\newacronym{fpr}{FPR}{False Positive Rate}
\newacronym{mm}{3DMM}{3D Morphable Model}
\newacronym{pca}{PCA}{Principal Component Analysis}
\newacronym{3ddfav3}{3DDFA-V3}{3D Dense Face Alignment Version 3}
\newacronym{sota}{SOTA}{state-of-the-art}
\newacronym{sgd}{SGD}{Stochastic Gradient Descent}
\newacronym{BA}{BA}{Balanced Accuracy}
\newacronym{BApoi}{$\mathrm{BA}_{\mathrm{poi}}$}{Balanced Accuracy}
\newacronym{AUCpoi}{$\mathrm{AUC}_{\mathrm{poi}}$}{AUC}
\newacronym{mlp}{MLP}{Multi-Layer Perceptron}
\newacronym{vit}{ViT}{Vision Transformer}
\newacronym{cls}{CLS}{classification}
\newacronym{clip}{CLIP}{Contrastive Language Image Pretraining}
\newacronym{poi}{POI}{Person-of-Interest}
\newacronym{ar}{AR}{Augmented Reality}
\newacronym{vr}{VR}{Virtual Reality}
\newacronym{mae}{MAE}{Masked Autoencoder}
\newacronym{ssl}{SSL}{Self-Supervised Learning}
\newacronym{vgg}{VGG}{Visual Geometry Group}
\newacronym{ntx}{NT-Xent}{Normalized Temperature-scaled Cross-Entropy}
\newacronym{auc}{AUC}{Area Under the Curve}
\newacronym{hq}{HQ}{Higher Quality}
\newacronym{lq}{LQ}{Lower Quality}
\newacronym{bamax}{$\mathrm{BA}$}{Balanced Accuracy}
\newacronym{ba0}{$\textrm{B-ACC}_{@ \textrm{thr}=0}$}{Balanced Accuracy at threshold 0}
\newacronym{tpr0}{$\textrm{TPR}_{@ \textrm{thr}=0}$}{True Positive Rate at threshold 0}
\newacronym{tprtheta}{$\textrm{TPR}(\theta)$}{True Positive Rate at threshold $\theta$}
\newacronym{fpr0}{$\textrm{FPR}_{@ \textrm{thr}=0}$}{False Positive Rate at threshold 0}
\newacronym{lpip}{LPIPS}{Learned Perceptual Image Patch Similarity}

\newacronym{cupid}{CUPID}{reConstructing UV texture maps for Person-of-Interest Deepfake detection}

\def\x{\mathbf{x}}

\def\I{\mathbf{I}}
\def\Iuv{\mathbf{I}_{\textrm{UV}}}
\def\Iuvhat{\hat{\mathbf{I}}_{\textrm{UV}}}

\def\params{{\boldsymbol{\Theta}}}

\def\Lrec{\mathcal{L}_{\text{rec}}}
\def\Lperc{\mathcal{L}_{\text{perc}}}
\def\Lcont{\mathcal{L}_{\text{cont}}}
\def\tokentesti{\mathbf{Z}^{\text{T}}_{i}}
\def\tokentestj{\mathbf{Z}^{\text{T}}_{j}}
\def\tokenrefi{\mathbf{Z}^{\text{R}}_{i}}
\def\tokenrefj{\mathbf{Z}^{\text{R}}_{j}}
\def\clstokentesti{\mathbf{z}^{\text{T}}_{i}}
\def\clstokentestj{\mathbf{z}^{\text{T}}_{j}}
\def\clstokenrefi{\mathbf{z}^{\text{R}}_{i}}
\makeatother

\glsdisablehyper

\def\BibTeX{{\rm B\kern-.05em{\sc i\kern-.025em b}\kern-.08em
    T\kern-.1667em\lower.7ex\hbox{E}\kern-.125emX}}
\begin{document}

\title{\glsentryshort{cupid}: Reconstructing UV Texture Maps for Interpretable Person-of-Interest Deepfake Detection}

\author{Giovanni Affatato \IEEEmembership{Student Member, IEEE}, Sara Mandelli \IEEEmembership{Member, IEEE}, Edoardo Daniele Cannas \IEEEmembership{Member, IEEE}, Paolo Bestagini \IEEEmembership{Member, IEEE}, and Stefano Tubaro \IEEEmembership{Senior Member, IEEE}
\thanks{The authors are with the Dipartimento di Elettronica, Informazione e Bioingegneria (DEIB), Politecnico di Milano, 20133 Milan, Italy.
This work was supported by the FOSTERER project, funded by the Italian Ministry of Education, University, and Research within the PRIN 2022 program.
This work was partially supported by the European Union - Next Generation EU under the Italian National Recovery and Resilience Plan (NRRP), Mission 4, Component 2, Investment 1.3, CUP D43C22003080001, partnership on ``Telecommunications of the Future'' (PE00000001 - program ``RESTART'') and by the Investment 1.3, CUP D43C22003050001, partnership on ``SEcurity and RIghts in the CyberSpace'' (PE00000014 - program ``FF4ALL-SERICS'').}}

\maketitle

\AddToShipoutPictureFG*{%
  \AtTextLowerLeft{\raisebox{-0.32in}{\parbox{\textwidth}{\centering\footnotesize
  This work has been submitted to the IEEE for possible publication. Copyright may be transferred without notice, after which this version may no longer be accessible.}}}%
}

\begin{abstract}
Deepfakes targeting a high-profile individual, known as \gls{poi}, are a threat to modern democracies and societies.
Current \gls{poi} deepfake detection methods still struggle to combine robustness to post-processing, efficiency and interpretability, key aspects of modern deepfake detectors.
In this paper we propose \glsentryshort{cupid}, a \gls{poi} video deepfake detector that combines UV texture maps, a facial appearance representation derived from 3D face reconstructions, with the representation learning capabilities of the \gls{mae}.

Our method does not require any deepfake videos in its training phase. Moreover, it does not even require including a specific \gls{poi} in the training set: the combination of UV texture maps extracted from real video frames and the \gls{mae} context-guided reconstruction yields a latent space that captures rich and discriminative facial features even for identities unseen during training.
In the testing phase, the embeddings extracted from a query video depicting the \gls{poi} can be matched against pristine reference videos
to assess the video authenticity.
Furthermore, operating in the UV space naturally provides an additional layer of interpretability. Specifically, we can extract decoded residual maps that highlight which facial regions of a test video deviate most from the identity representation of the corresponding \gls{poi}.

Experiments on four deepfake datasets show that \glsentryshort{cupid} outperforms
the current state of the art on most datasets and achieves the best overall robustness against strong downscaling and compression, while also providing substantially faster inference.
Our experimental code will be released at \href{https://github.com/polimi-ispl/CUPID}{polimi-ispl/CUPID}.
\end{abstract}
\vspace{-5pt}

\glsresetall


\section{Introduction}
\label{sec:introduction}
In recent years, the rapid advancements in the field of \gls{gai} have enabled the creation
of synthetic videos, popularly known as deepfakes, of unprecedented realism~\cite{koutlisVideoDeepfakeDetection2026}.
The malicious use of \gls{gai} technologies has fueled the spread of deepfakes targeting specific individuals,
such as politicians~\cite{allynDeepfakeVideoZelenskyy2022}, celebrities~\cite{AIgeneratedAdsUsing2024}, and high-profile corporate employees~\cite{magramoBritishEngineeringGiant2024}, with the aim of spreading misinformation and enabling fraudulent activities.

In response, the forensics community has developed specialized methods to protect a specific individual, referred to as the \gls{poi},
by exploiting the large amount of pristine data available for that person, such as videos of interviews, speeches, and public appearances~\cite{agarwalDetectingDeepFakeVideos2020, dongProtectingCelebritiesDeepFake2022,  agarwalProtectingWorldLeaders, agarwalWatchThoseWords2023, cozzolinoIDRevealIdentityawareDeepFake2021, cozzolinoAudioVisualPersonofInterestDeepFake2023, salvi2025phoneme}.
\gls{poi}-specific methods can be categorized into two main groups: methods that use the \gls{poi} data during training, and methods that use the \gls{poi} data only during inference.

The first group focuses on training specialized models capable of representing characteristic features of the \gls{poi},
such as facial appearance \cite{agarwalDetectingDeepFakeVideos2020,dongProtectingCelebritiesDeepFake2022}, facial expressions and movements \cite{agarwalProtectingWorldLeaders},
and audio-visual phoneme dynamics \cite{ agarwalWatchThoseWords2023}.
The main disadvantage of these methods is that they require training a separate detection model for each \gls{poi}, which in turn requires a large amount of data for that specific \gls{poi}.

\begin{figure}[t]
\centering
\includegraphics[width=.9\columnwidth]{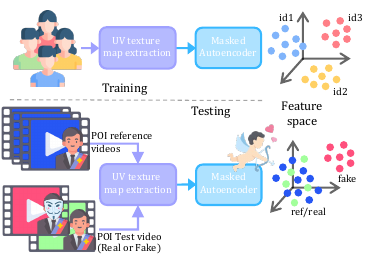}
\vspace{-10pt}
\caption{Overview of the proposed \glsentryshort{cupid} method for \gls{poi} deepfake detection. During training (top), a masked autoencoder encodes UV texture maps from real videos of many subjects into a general latent representation of facial identity. During inference (bottom), the same encoder maps pristine videos of the \gls{poi} into this latent space to build a reference identity distribution. A test video, whether real or manipulated, is then encoded and compared against this distribution for authenticity assessment.}
\label{fig:teaser}
\vspace{-15pt}
\end{figure}

The second group, in contrast, learns a general identity representation from real videos of many different subjects.
At inference time, these detectors are fed a reference set of pristine videos of the \gls{poi} and a test video portraying the \gls{poi}; detection is then performed by measuring the similarity, in the latent space, between the test video and the reference set.
Representative examples include ID-Reveal~\cite{cozzolinoIDRevealIdentityawareDeepFake2021},
which works on the visual-only domain (i.e., using the video frames and not the speech),
and POI-Forensics~\cite{cozzolinoAudioVisualPersonofInterestDeepFake2023}, which extends to the audio-visual domain by learning a joint embedding from face crops and audio spectrograms.
However, these methods do not fully address the problem of designing a \gls{poi} detector that is simultaneously robust to post-processing, computationally efficient, interpretable, and stable to threshold calibration.

In this paper, we propose \gls{cupid}, a novel visual-only \gls{poi} deepfake detector, whose overall pipeline is sketched in Fig.~\ref{fig:teaser}. The key idea
is that we do not operate directly on video frames but describe facial appearance in the UV texture space~\cite{egger3DMorphableFace2020}. UV texture maps, which we obtain by fitting a \gls{mm} to selected frames, project facial appearance onto a common image space, where corresponding pixels consistently represent the same facial region across identities and video frames~\cite{egger3DMorphableFace2020}.
As we show throughout the paper, this dense semantic correspondence provides a structured input representation that drives the robustness, efficiency, and interpretability of our detector.

From UV texture maps extracted from real videos of many subjects, we are able to learn a general latent space of facial identity by using a \gls{mae}~\cite{heMaskedAutoencodersAre2022}.
We further shape this space with a contrastive objective that pulls together representations of the same subject and pushes apart those of different subjects, yielding identity-discriminative features well suited to reference-set matching~\cite{chenSimpleFrameworkContrastive2020, wangMultiSimilarityLossGeneral2019}.

\gls{cupid} requires no \gls{poi} data in training, as the resulting space generalizes to unseen identities.
The \gls{poi} enters only at inference, where the \gls{mae} encodes UV texture maps derived from a set of pristine reference videos into a latent distribution that represents the subject's genuine appearance. Given a test video depicting the \gls{poi}, authenticity is assessed by how far its UV texture maps fall from this reference distribution.

Our main contributions are as follows:
\begin{itemize}
    \item We introduce \gls{cupid}, a \gls{poi}-specific video deepfake detector that combines UV texture maps with a \gls{mae} trained only on real videos.
    The method learns a general identity-aware representation and can be applied to unseen \glspl{poi} at inference time.
    \item We propose an interpretability procedure that leverages the structured nature of UV texture maps and the learned latent representation of the \gls{mae} to highlight the facial regions driving the detector's decisions. The resulting subject-level residual maps provide intuitive visual explanations for both real and manipulated videos.
    \item We compare our method against a \gls{poi}-specific baseline and general state-of-the-art
    deepfake detectors, providing an extensive evaluation on four deepfake datasets under both high-quality and low-quality settings, such as strong compression and resizing.
\end{itemize}

Our results show that \gls{cupid} outperforms the state of the art on most datasets, achieving the best overall robustness and substantially faster inference.
Moreover, the proposed methodology exhibits limited sensitivity to threshold calibration.

\section{Background}
\label{sec:background}

\subsection{3D Morphable Models and UV Texture Maps}
\label{subsec:back_3dmm}
\glsreset{mm}

\glspl{mm} are widely used in computer vision and computer graphics as generative models for facial geometry and appearance \cite{egger3DMorphableFace2020}.
A core requirement is that the 3D scans of human faces are brought into a dense vertex-by-vertex correspondence, meaning that each vertex index refers to the same semantic facial location across subjects.
For instance, if the $i$-th vertex identifies the nose tip in one scan, it represents the same anatomical point in all the others.

\begin{figure}[t]
  \centering
  \includegraphics[width=\columnwidth]{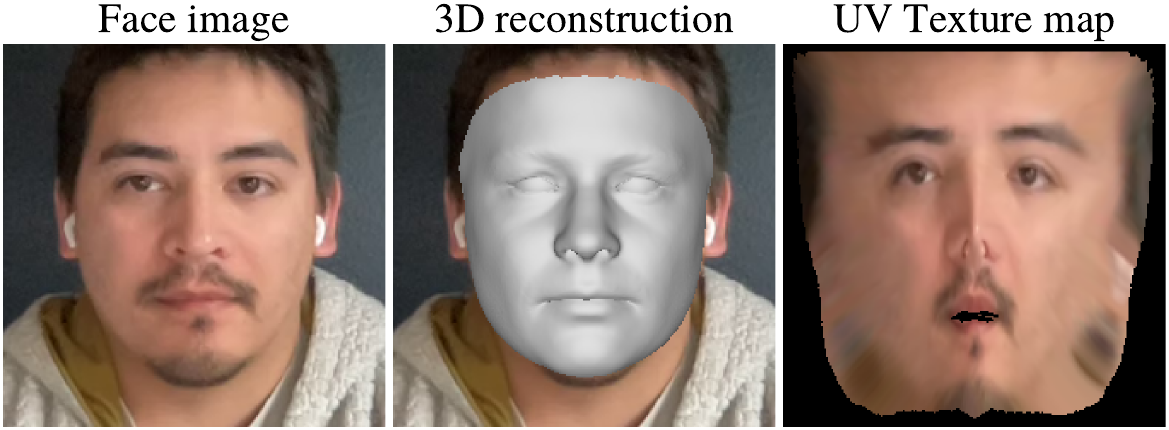}
  \caption{Example of 3D facial reconstruction and UV texture map extraction allowed by \glspl{mm}. From left to right: the input facial image, its 3D facial reconstruction (shown as a shaded rendering of the recovered surface projected onto the image plane)
  and the UV texture map.}
  \label{fig:mm_dec}
\vspace{-10pt}
\end{figure}

Let $\mathcal{S} = \{\mathbf{S}_1, \mathbf{S}_2, ..., \mathbf{S}_{N_\text{e}}\}$ be a collection of $N_\text{e}$ exemplar 3D faces.
Each sample $\mathbf{S}_i \in \mathcal{S}$ is represented by a matrix stacking the coordinates of its $N_\text{v}$ vertices, ordered consistently with the dense correspondence introduced above, so that the $j$-th column refers to the same facial point in every face: $\mathbf{S}_i = [\mathbf{v}_1^i, ..., \mathbf{v}_{N_\text{v}}^i] \in \mathbb{R}^{3 \times N_\text{v}}$, where $\mathbf{v}_j^i \in \mathbb{R}^3$ denotes the 3D coordinates of the $j$-th vertex.
Thanks to this shared ordering, a novel face
can be synthesized as a vertex-wise linear combination of the exemplar faces:
\begin{equation}
    \label{ed:lincomb}
    \mathbf{S}(\boldsymbol{\alpha}) = \mathbf{S}_1 \alpha_1 + \mathbf{S}_2 \alpha_2 + ... + \mathbf{S}_{N_\text{e}} \alpha_{N_\text{e}},
\end{equation}
where $\boldsymbol{\alpha} = [\alpha_1, \alpha_2, ... \alpha_{N_\text{e}}]^{\top}$ denotes the coefficients associated with the faces in $\mathcal{S}$.
Since the combination is built from realistic facial examples, \eqref{ed:lincomb} defines a plausible shape space that captures the spatial statistics of the vertex coordinates.

This geometric prior makes \glspl{mm} particularly useful for recovering a 3D face from a single 2D image.
Given a 3D shape $\mathbf{S}(\boldsymbol{\alpha}) \in \mathbb{R}^{3 \times N_\text{v}}$, its projection onto the image plane can be obtained by applying a known orthographic projection $\mathbf{T} \in \mathbb{R}^{2 \times 3}$ together with a similarity transformation, parameterized by a rotation matrix $\mathbf{R} \in \mathbb{R}^{3 \times 3}$, a scale factor $s$ and a 2D translation $\mathbf{t} \in \mathbb{R}^2$:
\begin{equation}
    \label{eq:3d_face_reconstruction}
    \mathbf{S}_{\textrm{2D}} = s \mathbf{T} \mathbf{R} \mathbf{S}(\boldsymbol{\alpha}) + \mathbf{t},
\end{equation}
with $\mathbf{S}_{\textrm{2D}} \in \mathbb{R}^{2 \times N_\text{v}}$.
In this setting, the 3D face reconstruction corresponds to estimating the parameters $\params = [s, \mathbf{R}, \mathbf{t},  \boldsymbol{\alpha}]$ from the observed 2D image.
An example of reconstructed face, visualized as a shaded rendering of the reconstructed 3D surface (projected onto the image plane via $\mathbf{S}_{\textrm{2D}}$), is shown in the second column of Fig.~\ref{fig:mm_dec}.

Another operation enabled by \glspl{mm} is UV texture map extraction~\cite{egger3DMorphableFace2020}.
In this regard, a UV texture map is an image that stores the color information of each vertex of the 3D face $\mathbf{S}(\boldsymbol{\alpha})$. The letters ``U'' and ``V'' denote the axes of the 2D texture map, to avoid confusion with X, Y, Z, the coordinates of the 3D space.
It is worth noting that this color information also corresponds to the color information of each point of the 2D face projection $\mathbf{S}_{\textrm{2D}}$~\cite{cosker2011facs}.

Once the parameters $\params$ have been estimated, an input facial image $\I$ can be sampled through a mapping $\mathrm{UV}: \mathbb{R}^2 \rightarrow \mathbb{R}^{2}$ that links each texture coordinate $\mathbf{p}_{\textrm{UV}} = (u, v)$ to its corresponding 2D point $\mathbf{p}_{\textrm{2D}} = (x, y)$ in the columns of $\mathbf{S}_{\textrm{2D}}$, such that $\mathrm{UV}(\mathbf{p}_{\textrm{UV}}) = \mathbf{p}_{\textrm{2D}}$.
Specifically, each pixel in the UV texture map $\mathbf{I}_{\textrm{UV}}$ of the image $\I$ is defined as
\begin{equation}
    \label{eq:uv_mapping}
    [\Iuv]_{u, v} = [\I]_{x, y}.
\end{equation}

\begin{figure}[t]
  \centering
  \includegraphics[width=\columnwidth]{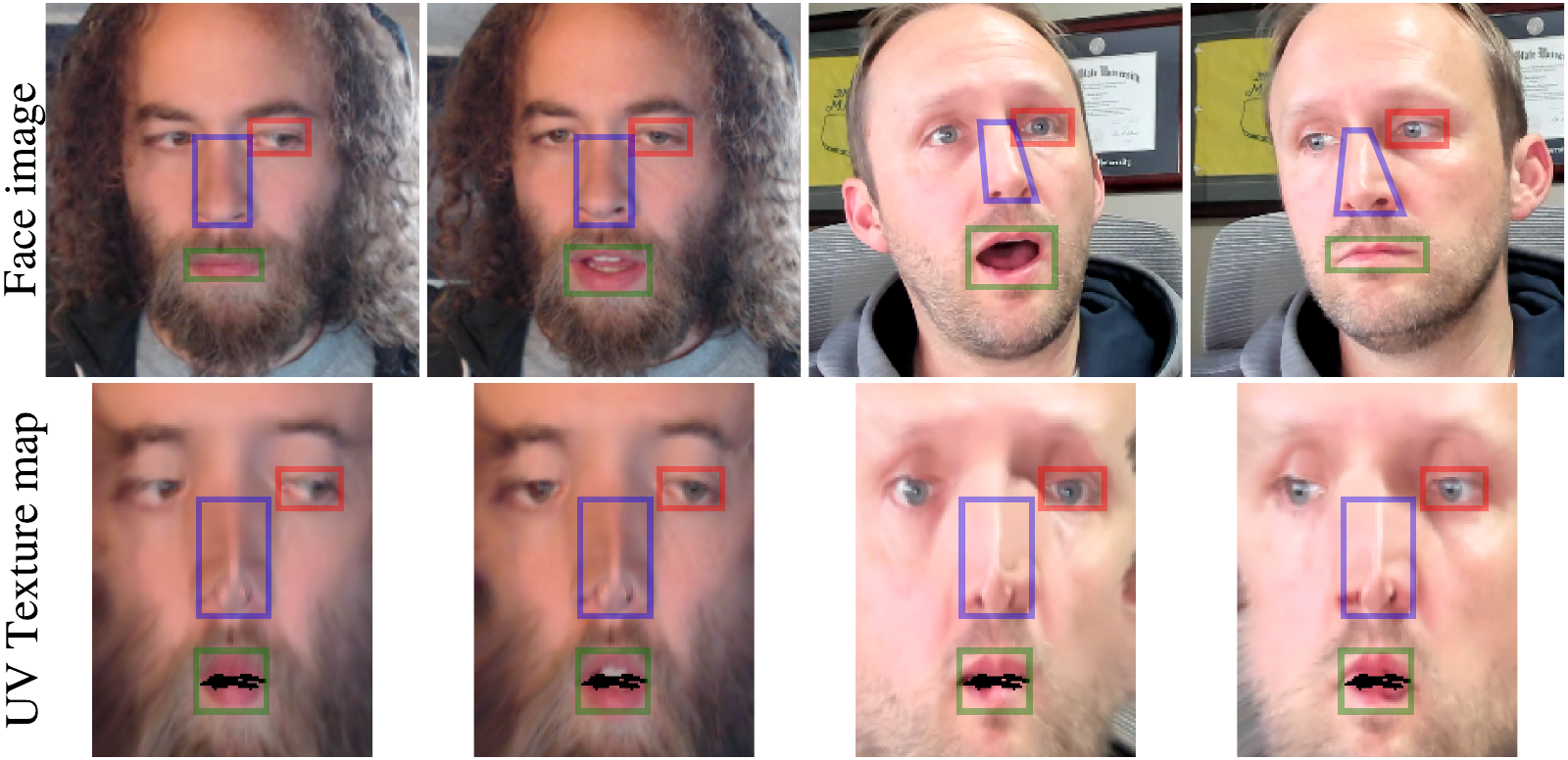}
  \caption{Face representations: facial images vs UV texture maps (center crop).
  We highlight the regions of the right eye, nose, and mouth for each representation. Notice how specific regions of the UV texture maps represent the same facial element independently of the pose and the subject.}
  \label{fig:mm_examples}
\end{figure}

An example of the resulting UV texture map is reported in the last column of Fig.~\ref{fig:mm_dec}.
A key property of this representation is that semantic correspondence is preserved in the UV space: the same pixel location corresponds to the same facial region across different subjects.
As a consequence, landmarks such as the nose tip, eyes, or mouth are mapped to consistent UV coordinates regardless of identity, expression, or pose.
Fig.~\ref{fig:mm_examples} shows some examples of this property.

\subsection{Masked Autoencoder}
\label{subsec:back_mae}

The \gls{mae} is a \gls{ssl} framework for visual representation learning introduced in \cite{heMaskedAutoencodersAre2022}.
Its objective is to reconstruct missing portions of an image from a partial observation, thereby forcing the model to capture the global structure and semantics of the input rather than relying only on local pixel correlations.
In practice, \gls{mae} is typically instantiated on top of a \gls{vit} \cite{dosovitskiyImageWorth16x162021a}, which naturally represents an image as a sequence of patches.

Let a generic image $\mathbf{X} \in \mathbb{R}^{H \times W \times 3}$ be partitioned into $N_\text{p}$ non-overlapping square patches, and let $\x = [\x_1, \x_2, ..., \x_{N_\text{p}}]$ denote the corresponding patch sequence.
A random masking procedure selects a subset $\mathcal{V} \subseteq \{1, \dots, {N_\text{p}}\}$ of visible patch indices and a complementary subset $\mathcal{M} = \{1, \dots, N_\text{p}\} \setminus \mathcal{V}$ of masked patch indices.
The masking ratio is defined as $\rho = |\mathcal{M}| / {N_\text{p}}$.
A key design choice of the \gls{mae} is the use of high masking ratios, making image reconstruction a non-trivial task that requires contextual reasoning over the visible content. This encourages the latent representation to capture dependencies between distant yet semantically related regions.

The architecture follows an asymmetric encoder-decoder design.
The encoder $E(\cdot)$ receives only the visible patches $\x_i, \; i   \in \mathcal{V}$, which are first linearly embedded into tokens and combined with positional encodings.
The resulting latent representation
\begin{equation}
    \mathbf{z} = E(\{\x_i\}_{i \in {\mathcal{V}}})
\end{equation}
is then passed to a lightweight decoder $D(\cdot)$, which re-introduces learned mask tokens at the missing positions and predicts the full patch sequence
\begin{equation}
    \hat{\x} = D(\mathbf{z}, \mathcal{M}).
\end{equation}

Training is performed by minimizing a reconstruction loss between the original and the predicted patch content, computed only on the patches belonging to the $\mathcal{M}$ set.
After training, the decoder is discarded and the encoder is retained as a feature extractor for many downstream tasks.

\section{Proposed Methodology}
\label{sec:method}

\begin{figure*}[t]
	\centering
	\includegraphics[width=\textwidth]{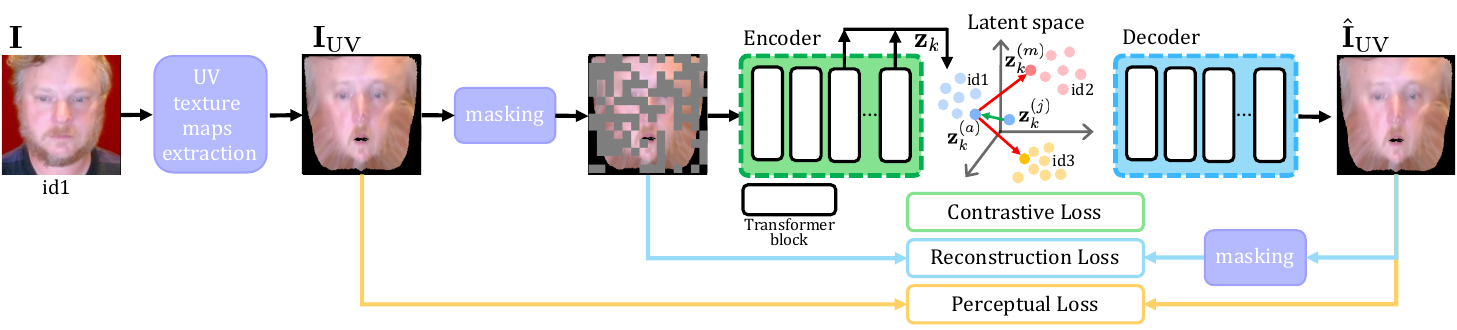}
    \vspace{-18pt}
	\caption{Overview of the \gls{cupid} training pipeline. For each frame of a real video of a known identity,
    a face crop is fed to the \gls{mm}-based UV extractor, which projects the facial appearance onto the UV space, yielding one UV texture map $\Iuv$ per frame.
    Each UV texture map is partitioned into non-overlapping patches and $50\%$ of them are randomly masked.
    Patches
    are then linearly embedded, augmented with positional encodings and a learnable \glsentryshort{cls} token.
	The \gls{vit} encoder processes only the visible tokens, while the decoder reconstructs the full UV texture map $\Iuvhat$.
    Training minimizes a joint objective that combines a masked reconstruction loss,
    a perceptual loss and a multi-layer contrastive loss.
    }
	\label{fig:method}
\vspace{-10pt}
\end{figure*}

In this section, we first formalize the detection problem and overview our proposed methodology, then we describe the proposed training pipeline and illustrate the deployment stage. Finally, we show how the proposed UV-based \gls{mae} strategy enhances the interpretability of the method by highlighting the facial regions of the analyzed subject that contribute most to the detection process.

\subsection{\texorpdfstring{\gls{poi}}{POI} Deepfake Detection}
\label{subsec:method_problem}
We tackle \gls{poi} deepfake detection in an open-set scenario.
A target identity, the \gls{poi}, is represented by a set of pristine reference videos $\mathcal{D}_{\textrm{ref}} = \{V_1, \dots, V_{N_{\text{vid}}}\}$, all genuinely portraying that subject.
Given a test video $V$ attributed to the same \gls{poi}, the task is to detect whether $V$ is authentic or a deepfake (i.e., to assign a label $y \in \{\textrm{real}, \textrm{fake}\}$) based on $\mathcal{D}_{\textrm{ref}}$.

Following the work of Cozzolino et al.~\cite{cozzolinoAudioVisualPersonofInterestDeepFake2023}, we cast detection as the thresholding of an identity-similarity score.
Let $E$ be an encoder that, applied to each sampled frame, maps a video into a set of frame-level identity descriptors. We score a test video by how closely its descriptors $E(V)$ match those of the reference videos, $\{E(V_i)\}_{i=1}^{N_{\text{vid}}}$, through a similarity function $S(V, \mathcal{D}_{\textrm{ref}}) \in \mathbb{R}$.
Detection is then performed as
\begin{equation}
\label{eq:decision_rule}
\hat{y} =
\begin{cases}
\textrm{real}, & \text{if } S(V, \mathcal{D}_{\textrm{ref}}) \ge \gamma, \\
\textrm{fake}, & \text{otherwise},
\end{cases}
\end{equation}
where $\gamma$ is a decision threshold.

We consider a scenario shaped by two constraints, chosen to be both realistic and challenging:
\begin{itemize}
    \item \emph{Open-set manipulations}: the forgery techniques used to produce deepfakes are unknown at training time.
    \item \emph{\gls{poi}-agnostic detection}: no data of the target \gls{poi} is available during training, so the encoder cannot be specialized to any specific identity.
\end{itemize}

Under these constraints, the problem reduces to learning the encoder $E$ such that the induced similarity $S(\cdot, \cdot)$ separates genuine videos of a \gls{poi} from impersonations, while generalizing to identities and manipulations unseen during training.
Section~\ref{subsec:method_training} describes how $E$ is trained on real data alone, while Section~\ref{subsec:method_deployment} instantiates a pairwise similarity test, yielding a final score that is compared against $\gamma$.

\subsection{Training pipeline}
\label{subsec:method_training}
During training, the model is encouraged to learn latent representations that are highly descriptive of subjects' identity, enabling it to subsequently discriminate between real and fake versions of the same individual.
The training pipeline (illustrated in Fig.~\ref{fig:method}) consists of two main stages: UV texture map extraction and \gls{mae} encoding-decoding.
In the following, we provide further details on each stage.

\subsubsection{UV Texture Map Extraction}
\label{subsubsec:method_uv_extraction}
Given a video of a specific identity, we uniformly sample $N_{\text{frm}}$ frames and extract an aligned face crop from each of them.
The aligned facial image $\I$ is then processed by the \gls{mm}-based UV extractor, which fits the face model and projects the facial appearance onto the canonical UV coordinate space, producing one UV texture map $\Iuv$ per sampled frame (see~\eqref{eq:uv_mapping}).
The use of UV texture maps instead of raw video frames represents one of the main differences between our approach and the state of the art.
As discussed in Section~\ref{sec:background}, this representation enables the disentanglement of identity-related appearance from pose and expression variations, thereby providing a standardized input that facilitates feature learning.
Furthermore, previous works have shown that UV texture maps are more robust than raw video frames to various processing operations~\cite{affatato3DMorphableModels2025}.

\subsubsection{MAE encoding-decoding}
\label{subsubsec:method_mae}

Building on the standard \gls{mae} formulation described in Section~\ref{sec:background}, we propose a \gls{vit}-based encoder-decoder operating on UV texture maps.
The input UV texture map $\Iuv$ is partitioned into non-overlapping patches, embedded into $C$-dimensional tokens, augmented with learnable positional embeddings and prepended with a learnable \gls{cls} token. During training, a fraction $\rho$ of the patches is randomly masked.
The encoder processes only the visible tokens, while the decoder reconstructs $\Iuvhat$, inserting learned mask tokens at the missing positions.

We train the \gls{mae} by combining three complementary loss components to jointly encourage accurate masked reconstruction, perceptually coherent facial representations and identity-aware latent embeddings.
Specifically, we impose a joint objective including these loss components:
\begin{itemize}
    \item A masked reconstruction loss, i.e., a per-pixel \gls{mse} computed on the masked patches. This encourages the model to learn rich facial texture representations by reconstructing the masked content; we define this loss as $\Lrec$.
    \item A multi-layer contrastive loss that enforces identity clustering in the latent space across multiple encoder depths; we define this loss as $\Lcont$.
    \item A perceptual loss that, rather than comparing raw pixels, matches the deep feature maps of a pretrained network to promote perceptually coherent, artifact-free reconstructions; we define this loss as $\Lperc$.
\end{itemize}

The total loss is a weighted combination of three components:
\begin{equation}
\mathcal{L} = \tau \cdot \Lrec  + (1 - \tau) \cdot \Lcont + \lambda \cdot \Lperc,
\end{equation}
where $\tau$ balances the reconstruction and contrastive objectives, while $\lambda$ weights the perceptual term.
In particular, $\tau$ controls the trade-off between the two main learning objectives that shape the latent space, namely masked reconstruction and identity-aware contrastive supervision, whereas $\lambda$ only regulates an auxiliary perceptual refinement term that mitigates the grid artifacts typically produced by \gls{mae}-only reconstruction~\cite{heMaskedAutoencodersAre2022}.
We provide more details on each loss term in the following paragraphs.

\paragraph{Masked Reconstruction Loss}
For this term, we follow the original \gls{mae} formulation~\cite{heMaskedAutoencodersAre2022} and
measure the per-pixel \gls{mse} only on the masked patches.
Accordingly, the masked reconstruction loss is defined as:

\begin{equation}
\Lrec = \frac{1}{|\mathcal{M}|} \sum_{p \in \mathcal{M}} \| [\Iuvhat]_p - [\Iuv]_p \|_\textrm{F}^2,
\end{equation}
where $\mathcal{M}$ denotes the set of masked patch positions and $[\Iuv]_p, [\Iuvhat]_p$ are the original and reconstructed pixel values of patch $p$.
$\|\cdot\|_\textrm{F}$ denotes the Frobenius norm.

\paragraph{Contrastive Loss}

The contrastive component enforces identity-discriminative representations across multiple depths of the encoder. Let $\mathcal{K}$ denote the set of selected intermediate encoder layers, and let $\mathbf{z}_k \in \mathbb{R}^{C}$ be the unit-norm \gls{cls} embedding extracted from layer $k$.
For each $k \in \mathcal{K}$, we apply the \gls{ntx} \cite{chenSimpleFrameworkContrastive2020} loss.
In practice, given a batch of samples with different identity labels, we denote by $\mathcal{B}$ the set of sample indices in the batch, by $a \in \mathcal{B}$ the index of an anchor sample, and define the set of positive samples $\mathcal{P}(a)$ as all the other samples in the batch sharing the same identity.
The \gls{ntx} loss for anchor $a$ at layer $k$ can be defined as:

\begin{equation}
\mathcal{L}_{\text{NT-Xent}}^{(k,a)} = -\frac{1}{|\mathcal{P}(a)|} \sum_{j \in \mathcal{P}(a)} \log \frac{\exp(\mathbf{z}_k^{(a)} \cdot \mathbf{z}_k^{(j)} / t)}{\sum\limits_{m \in \mathcal{B} \setminus \{a\}} \exp(\mathbf{z}_k^{(a)} \cdot \mathbf{z}_k^{(m)} / t)},
\end{equation}
where $t$ is the temperature parameter.
The total contrastive loss aggregates over all anchors $a \in \mathcal{B}$ and combines the selected layers through a weighted sum:

\begin{equation}
\Lcont = \sum_{a \in \mathcal{B}} \sum_{k \in \mathcal{K}} w_k \cdot \mathcal{L}_{\text{NT-Xent}}^{(k,a)},
\end{equation}
where the layer-specific weights $w_k$ balance the contribution of the selected encoder depths, with deeper layers typically capturing higher-level semantics \cite{yueUnderstandingMaskedAutoencoders2023}.

The formulation above involves every positive and negative pair available in the batch. In practice, to improve the efficiency of contrastive learning, we employ a Multi-Similarity Miner~\cite{wangMultiSimilarityLossGeneral2019}, applied independently at each selected layer $k \in \mathcal{K}$, which restricts the loss computation to the most informative comparisons.
Relying on the cosine-similarity relations among the layer's \gls{cls} embeddings $\mathbf{z}_k$ within the batch, the miner retains only the hard samples, namely positive pairs with relatively low cosine similarity and negative pairs with relatively high cosine similarity.

\paragraph{Perceptual Loss}

Reconstructing the masked patches with a purely pixel-wise \gls{mse} objective tends to produce blocky, grid-like artifacts at the patch boundaries, a known limitation of \gls{mae} reconstructions~\cite{heMaskedAutoencodersAre2022}.
To suppress these artifacts and obtain smoother, perceptually coherent reconstructions, we complement the pixel-wise term with a perceptual loss~\cite{johnsonPerceptualLossesRealTime2016} computed on the feature maps of a pretrained \gls{vgg}-16 network~\cite{simonyanVeryDeepConvolutional2015}.
Concretely, the reconstructed UV texture map $\Iuvhat$ and the corresponding target UV texture map $\Iuv$ are fed separately to the frozen \gls{vgg}-16, and feature maps are extracted from a set of intermediate layers $\mathcal{F}$ for both forward passes.
The perceptual loss is the mean L1 distance across these feature maps:

\begin{equation}
\Lperc = \frac{1}{|\mathcal{F}|} \sum_{l \in \mathcal{F}} \| \phi_l(\Iuvhat) - \phi_l(\Iuv) \|_1,
\end{equation}
where $\phi_l(\cdot)$ denotes the feature map at layer $l$ of the frozen \gls{vgg}-16 network and $\mathcal{F}$ is the set of selected perceptual layers.

\vspace{-10pt}
\subsection{Network Deployment}
\label{subsec:method_deployment}
The design of our architecture and the large amount of real data used in training generate a latent space that generalizes well even to subjects not seen during training.
This means that the specific \gls{poi} under investigation does not need to be enrolled in the training set.
Therefore, given a \gls{poi} to focus on, we can perform deepfake detection relying only on a small set of pristine videos of the subject.

\subsubsection{Embeddings Extraction}

For each test \gls{poi}, we assume access to the reference set $\mathcal{D}_{\textrm{ref}}$, comprising $N_{\text{vid}}$ pristine videos. From each reference video, we uniformly sample $N_{\text{frm}}$ frames; therefore, $\mathcal{D}_{\textrm{ref}}$ yields $N_\text{r} = N_{\text{vid}} N_{\text{frm}}$ frame-level samples.
Given an analyzed test video of the same \gls{poi}, we uniformly sample $N_\text{t}$ frames, obtaining the corresponding test set of frame-level samples.

At deployment stage, the full UV texture maps (without masking) extracted from each analyzed face are fed to the encoder, which produces last-layer token embeddings for each sampled frame.
For a given subject, we denote the set of reference last-layer token embeddings by $\mathcal{R} = \{\tokenrefi\}_{i=1}^{N_\text{r}}$ and the test last-layer token embeddings by $\mathcal{T} = \{\tokentesti\}_{i=1}^{N_\text{t}}$, with $\tokenrefi, \tokentesti \in \mathbb{R}^{M \times C}$, where $M$ is the number of tokens and $C$ is the token dimension.
The first token of each last-layer embedding is the \gls{cls} embedding; we define the sets of unit-norm \gls{cls} embeddings by $\mathcal{R}_{\mathrm{CLS}} = \{\clstokenrefi\}_{i=1}^{N_\text{r}}$ for the reference set and $\mathcal{T}_{\mathrm{CLS}} = \{\clstokentesti\}_{i=1}^{N_\text{t}}$ for the test set, with $\clstokenrefi, \clstokentesti \in \mathbb{R}^{C}$.

\subsubsection{Pairwise Similarity Test}
Our pairwise similarity test compares the test set $\mathcal{T}_{\mathrm{CLS}}$ against the reference set $\mathcal{R}_{\mathrm{CLS}}$.
We use the \gls{cls} embeddings for scoring because they provide a compact texture-level summary of the whole token sequence~\cite{heMaskedAutoencodersAre2022}, enabling direct and efficient similarity computation, while the full token embeddings are retained for the interpretability analysis.
The pairwise similarity between the $i$-th reference unit-norm \gls{cls} embedding and the $j$-th test unit-norm \gls{cls} embedding is then defined as

\begin{equation}
s_{ij} = \left\langle \clstokenrefi, \clstokentestj \right\rangle, \, \, i \in \{1, ..., N_\text{r}\}, \, j \in \{1, ..., N_\text{t}\},
\end{equation}
where $\left\langle \cdot , \cdot  \right\rangle$ denotes the inner product.
The final similarity score, i.e., the similarity function of~\eqref{eq:decision_rule}, is then defined as the maximum similarity over all reference--test pairs:

\begin{equation}
S(V, \mathcal{D}_{\textrm{ref}}) = \max_{i,\, j} s_{ij}.
\end{equation}

Our methodological choice reflects the operational assumption that a genuine test video should contain at least some test features whose embeddings remain close to the support of the \gls{poi}'s authentic reference distribution. These features are expected to yield high similarities to the reference set, whereas manipulated videos should produce consistently lower similarities.
\vspace{-10pt}

\subsection{Interpretable POI Deepfake Detection}
\label{subsec:method_interpretability-mm-mae}

\begin{figure}[t]
\centering
\includegraphics[width=0.75\columnwidth]{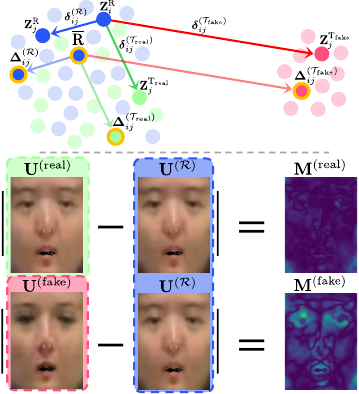}
\vspace{-5pt}
\caption{Sketch of the interpretability analysis enabled by \gls{cupid}.
Top: a latent-space view of the quantities involved, with the reference set in blue and the real and fake test sets in green and red, respectively.
Bottom: examples of decoded displacement maps and of the derived interpretability maps.
}
\label{fig:interp}
\vspace{-15pt}
\end{figure}

The synergy between the dense correspondence induced by \glspl{mm} and the rich latent space learned by the \gls{mae} allows us to extend our proposed methodology with an interpretability layer.
Building upon our previous work \cite{affatatoDecodingSyntheticFace2025}, our framework localizes in the UV space the facial regions most responsible for a fake decision, as sketched in Fig.~\ref{fig:interp}.

At a high level, our idea is to turn the latent space back into faces and ask which facial regions make a test video look unlike the genuine subject.
Working in the latent space means comparing faces in terms of the identity-bearing features on which the detection score itself is computed rather than in terms of raw pixels. In fact, pixel-level differences are often dominated by nuisance factors such as illumination, color, and \gls{mm}-fitting artifacts.

To this end, we first summarize the subject's genuine appearance into a latent prototype, which we define as the \emph{centroid} of its reference last-layer embeddings.
Then, each reference--test pair provides a displacement, which we apply to the centroid and decode, thus rendering the change that the test video induces on the prototype directly on the face.
The regions where the test-induced change exceeds the natural subject variability are those we deem responsible for a fake decision.

We define the reference centroid of a \gls{poi} subject as

\begin{equation}
\overline{\mathbf{R}} = \frac{1}{N_\text{r}}\sum_{i=1}^{N_\text{r}} \tokenrefi,
\end{equation}

which represents the canonical appearance of the subject in the latent space: averaging over multiple references suppresses sample-specific variation (pose and expression residuals, lighting) while preserving identity-related structure.
Averaging is meaningful because \glspl{mm} align all faces onto a canonical UV space: token positions, and the pixels they decode to, correspond to the same facial region across subjects and poses.

Given a test video of the subject, we compute the pairwise latent displacement $\boldsymbol{\delta}_{ij}^{(\mathcal{T})} = \tokentestj - \tokenrefi$ for each pair $(\tokenrefi, \tokentestj)$ of reference and test last-layer
embeddings, i.e., $(i, j) \in \mathcal{P}_{\mathcal{T}} = \{1, ..., N_\text{r}\} \times \{1, ..., N_\text{t}\}$.
This represents the latent change that maps the reference appearance $i$ onto the test appearance $j$.
We repeat this procedure on reference pairs alone, defining $\boldsymbol{\delta}_{ij}^{(\mathcal{R})} = \tokenrefj - \tokenrefi$ for $(i, j) \in \mathcal{P}_{\mathcal{R}} = \{(i, j) \in \{1, ..., N_\text{r}\} \times \{1, ..., N_\text{r}\} : i \neq j\}$. This shows how much the prototype varies naturally among genuine samples.
These two operations allow us to define a generic \textit{pairwise latent displacement} $\boldsymbol{\delta}_{ij}^{(X)}$, for $X \in \{\mathcal{T}, \mathcal{R}\}$ and $(i,j) \in \mathcal{P}_X$.

Rather than decoding these displacements directly, we replay them from the subject centroid, constructing the \emph{centroid-anchored displacements}
\begin{equation}
\boldsymbol{\Delta}_{ij}^{(X)} = \overline{\mathbf{R}} + \boldsymbol{\delta}_{ij}^{(X)}.
\end{equation}
Anchoring is crucial for decoding, as the centroid supplies the same base appearance to every decoded point of both constructions: what is ultimately compared is a distribution of \emph{changes} rendered on the same prototype, rather than the absolute appearances of two different videos, whose difference would also reflect video-level nuisances such as illumination.
The reference-anchored points (i.e., $\boldsymbol{\Delta}_{ij}^{(\mathcal{R})}$) reasonably scatter around the centroid itself, while the test-anchored points (i.e., $\boldsymbol{\Delta}_{ij}^{(\mathcal{T})}$) scatter around the centroid shifted by the systematic reference-to-test discrepancy. For an authentic video, this discrepancy will likely be small and no systematic offset will separate the two clouds. In contrast, we conjecture that a manipulation displaces the center of the test-anchored cloud, and this is precisely the component our analysis visualizes (see Fig.~\ref{fig:interp} for exemplifying visualizations).

Let $D : \mathbb{R}^{M \times C} \to \mathbb{R}^{H \times W}$ denote the \gls{mae} decoder applied without masking.
We average the decoded centroid-anchored displacements as
\begin{equation}
\mathbf{U}^{(X)} = \frac{1}{|\mathcal{P}_X|} \sum_{(i,j) \in \mathcal{P}_X} D\!\left(\boldsymbol{\Delta}_{ij}^{(X)}\right).
\end{equation}
We average over all pairs because any single displacement risks confounding our analysis with pair-specific factors such as expression and illumination: aggregating many pairs attenuates this effect.
Note that, in case of $\mathbf{U}^{(\mathcal{T})}$, the construction operates on the same reference--test pairs scored by the similarity test of Section~\ref{subsec:method_deployment}, so the interpretability layer explains the very same comparisons that drive the detection score.

The interpretability map for the test video is then
\begin{equation}
\mathbf{M}^{(\mathcal{T})} = \left|\mathbf{U}^{(\mathcal{T})} - \mathbf{U}^{(\mathcal{R})}\right|.
\end{equation}
The reference-only term $\mathbf{U}^{(\mathcal{R})}$ acts as a subject-specific normalizer, so the map $\mathbf{M}^{(\mathcal{T})}$ highlights, pixelwise in the decoded UV space, the regions where the test-related reconstruction of the prototype \textit{departs from what is already explained} by intra-reference variability, such as facial characteristic variations, illumination residuals, or pose-dependent texture artifacts left by the \gls{mm} fitting.
Moreover, since both terms average the nonlinear decoder over point clouds built by the same construction, decoder-induced distortions affect them similarly and are largely removed by the subtraction, so that $\mathbf{M}^{(\mathcal{T})}$ mainly reflects the systematic discrepancy between the test and reference appearances.

Consistently with the latent-space picture above, $\mathbf{M}^{(\mathcal{T})}$ should remain small and spatially diffuse for a real input, whereas we expect that manipulated embeddings deviate systematically, concentrating the map on the affected facial regions: a manipulation altering only the mouth, for instance, should push the decoded prototype away from $\mathbf{U}^{(\mathcal{R})}$ precisely there, yielding a spatially localized signature of the forgery.
In Fig.~\ref{fig:interp} we show some examples of the generated maps.

\section{Experimental Setup}
\label{sec:experimental-setup}
\subsection{Datasets}
Since we are interested in the \gls{poi} setting, we focus on datasets that provide subject identity metadata, and not only binary real/fake labels.

\subsubsection{Training Dataset}
We train our \gls{mae} on VoxCeleb2 \cite{chungVoxCeleb2DeepSpeaker2018}, a large-scale audio-visual corpus containing 6,112 identities, 150,480 videos, and 1,128,246 utterances in total, with 5,994 identities in the training split and 118 in the validation split.
Videos consist of face-cropped frames of size $224 \times 224$ pixels.
To avoid any identity overlap with the evaluation data (see details below), we remove the 500 identities used to construct the FakeAVCeleb dataset~\cite{khalidFakeAVCelebNovelAudioVideo2022}.
After this filtering step, we balance the training set by sampling 21 videos per identity, where 21 is the minimum number of available videos among the retained identities.
This gives a uniform identity distribution during training and prevents the model from being dominated by highly represented subjects.

\subsubsection{Test Datasets}
We evaluate our method on four datasets that cover a wide range of deepfake generation settings, from controlled face-swapping benchmarks to more modern avatar-based synthesis pipelines.
We evaluate each test set in two quality settings: \gls{hq}, corresponding to its original format, and \gls{lq}, corresponding to an additional H.264-compressed version produced with \texttt{crf40}.
In addition, for datasets that include LipSync-generated samples, we exclude those videos from evaluation because LipSync mainly edits the mouth region while leaving the identity cues unchanged; this scenario does not match our target setting of full-identity manipulation.

\paragraph{DF-TIMIT}
DF-TIMIT was introduced in \cite{korshunovDeepFakesNewThreat2018} and built from VidTIMIT \cite{sandersonMultiRegionProbabilisticHistograms2009}.
VidTIMIT contains 43 subjects with 13 videos per subject, recorded in a controlled setting at a resolution of $512 \times 384$ pixels.
The deepfake subset was created by manually selecting 16 visually similar subject pairs, involving 32 identities overall, and then generating bidirectional face-swaps with an autoencoder-based pipeline, for a total of 320 fake videos and 416 real videos taken from VidTIMIT.

\paragraph{FakeAVCeleb}
FakeAVCeleb \cite{khalidFakeAVCelebNovelAudioVideo2022} was built from 500 real VoxCeleb2 videos, one per celebrity, and contains a total of 19,500 fake videos with a resolution of $224 \times 224$ pixels.
The fake content was generated with FaceSwap \cite{korshunovaFastFaceSwapUsing2017} and FSGAN \cite{nirkinFSGANSubjectAgnostic2019a} for face-swapping, SV2TTS \cite{jiaTransferLearningSpeaker2018} for voice cloning and Wav2Lip \cite{prajwalLipSyncExpert2020} for audio-driven facial reenactment.
The manipulated dataset is organized into real-audio/fake-video, fake-audio/fake-video, and fake-audio/real-video subsets.
In our experiments, we consider only the first two subsets, since our method targets video manipulations only; for each identity, the real samples are taken directly from VoxCeleb2 so that their number matches that of the corresponding fake videos.

\paragraph{KoDF}
KoDF \cite{kwonKoDFLargescaleKorean2021} contains 403 subjects, 62,166 real and 175,776 fake high-resolution ($1920 \times 1080$) clips.
Fake videos were generated with six synthesis models: FaceSwap \cite{korshunovaFastFaceSwapUsing2017}, DeepFaceLab \cite{liuDeepfacelabIntegratedFlexible2023}, FSGAN \cite{nirkinFSGANSubjectAgnostic2019a}, FOMM \cite{siarohinFirstOrderMotion2019}, ATFHP \cite{yiAudiodrivenTalkingFace}, and Wav2Lip \cite{prajwalLipSyncExpert2020}.
KoDF is one of the largest identity-labeled deepfake datasets and includes both face-swapping and face-reenactment manipulations.

\paragraph{DeepSpeak}
DeepSpeak \cite{barringtonDeepSpeakDataset2025} contains 16,043 real videos and 14,005 fake videos spanning 500 different identities, with a resolution of $1280 \times 720$ pixels.
A key property of DeepSpeak is that the fake samples were created with large-scale identity matching (i.e., swaps and reenactments are performed between visually similar subjects) and a wide range of recent generation tools: it includes 14 video synthesis and 3 voice-cloning engines overall.
Across its releases, these include face-swap methods such as FaceFusion \cite{FacefusionFacefusion2026}, INSwapper \cite{wangHaofanwangInswapper2026} and SimSwap \cite{chenSimSwapEfficientFramework2020}; LipSync methods such as Wav2Lip \cite{prajwalLipSyncExpert2020}, VideoRetalking \cite{chengVideoReTalkingAudiobasedLip2022}, Diff2Lip \cite{mukhopadhyayDiff2LipAudioConditioned2024} and LatentSync \cite{liLatentSyncTamingAudioConditioned2024}; avatar-style generators such as LivePortrait \cite{guoLivePortraitEfficientPortrait}, HelloMeme \cite{zhangHelloMemeIntegratingSpatial} and Memo \cite{zhengMEMOMemoryGuidedDiffusion}.
Because the dataset preserves participant identity metadata and covers recent synthesis pipelines, it provides a challenging benchmark for cross-dataset evaluation.

\vspace{-10pt}
\subsection{Evaluation Metrics}
We report two main metrics: the \gls{auc} of the \gls{roc} curve and the \gls{ba}.
\gls{auc} measures the ranking quality independently of the decision threshold.
Regarding \gls{ba}, we always consider the maximum balanced accuracy over all possible thresholds.
We evaluate these metrics in two different setups, which are equally important since they represent distinct realistic scenarios.

In a first setup, we report the dataset-level \gls{auc} and \gls{ba} by pooling all subjects of a dataset together and using a single threshold for the whole dataset. This setting is relevant when the goal is to deploy a general and scalable system that can operate on all \glspl{poi} a priori, without per-subject calibration.

In a second setup, we compute the \gls{auc} and \gls{ba} independently for each identity in the dataset and then report the arithmetic mean across subjects.
This setting allows the threshold to be optimized separately for each subject.
This is particularly relevant in the \gls{poi} setting, where the system is deployed for a \textit{specific target identity} and a small amount of identity-specific validation data can be used to tune the operating threshold. Therefore, this setup allows us to measure how well the method separates real and fake samples once adapted to the \gls{poi} of interest.

For the rest of the paper, we denote \gls{poi}-level metrics as $\mathrm{AUC}_{\mathrm{poi}}$ and $\mathrm{BA}_{\mathrm{poi}}$, while dataset-level metrics are denoted simply as $\mathrm{AUC}$ and $\mathrm{BA}$.
In summary, the former setup reflects the more restrictive and practically relevant scenario in which one operating point must be shared across all identities in the dataset, while the latter highlights subject-specific separability after \gls{poi}-specific calibration.

\vspace{-10pt}
\subsection{State of the Art Baselines}
In our experiments, we compare against publicly available deepfake detection methods and one \gls{poi}-specific baseline.

As general baselines, we consider RealForensics~\cite{haliassosLeveragingRealTalking2022b}, LipForensics~\cite{haliassosLipsDontLie2021a} and FTCN~\cite{zhengExploringTemporalCoherence2021}. We selected these methods because they are publicly available and provide pretrained weights that can be evaluated directly in our setting. All three models are trained on FaceForensics++~\cite{rosslerFaceForensicsLearningDetect2019a}, which does not overlap with the datasets used in our generalization analysis. This makes the comparison cleaner, since the reported results reflect cross-dataset generalization rather than adaptation to the target benchmarks.
We also include the Seferbekov model \cite{seferbekovSelimsefDfdc_deepfake_challenge2026}, which was trained on the DFDC dataset \cite{dolhanskyDeepFakeDetectionChallenge2020}, since it was the winner of the DFDC challenge.

As a \gls{poi}-specific reference, we evaluate POI-Forensics \cite{cozzolinoAudioVisualPersonofInterestDeepFake2023}, one of the newest \gls{poi}-specific methods and, to the best of our knowledge, the only one publicly available.
This method was trained on the full VoxCeleb2 dataset \cite{chungVoxCeleb2DeepSpeaker2018}, which is the closest available training setting to our task. It is important to note that POI-Forensics was designed as a multimodal framework, analyzing both audio and visual modalities. Nonetheless, it is possible to consider only the visual modality, which already provides accurate results. To provide a fair comparison with our proposed strategy, we compare with the video-only setup.

\vspace{-10pt}
\subsection{Architecture Details}
The UV texture map pipeline uses RetinaFace~\cite{dengRetinaFaceSingleShotMultiLevel2020} for face extraction and alignment, and \gls{3ddfav3}~\cite{wang3DFaceReconstruction2024a} for 3D face reconstruction.
We set the UV texture map resolution to $224 \times 224 \times 3$, consistent with the resolution of the training dataset VoxCeleb2 and with standard \gls{vit}/\gls{mae} practice.

We use non-overlapping patches of size $16 \times 16$, which yields $M = 197$ patch tokens per sample (including the \gls{cls} token).
Each patch is projected to a token dimension $C = 192$.
The encoder has $L = 12$ Transformer blocks with 3 attention heads per block, while the decoder uses 4 Transformer blocks with 3 attention heads and hidden dimension of 192.
The reconstruction head predicts a flattened patch of dimension $3 \times 16^2 = 768$ for each token.
This corresponds to reconstructing the three RGB channels over all $16 \times 16$ pixels of each patch.

\vspace{-10pt}
\subsection{Training Configuration}
\label{subsec:training_configuration}
Training hyperparameters were selected using the Bayesian sweep search provided by Weights \& Biases \cite{wandb2020}.
The mask ratio is set to $\rho = 0.5$, meaning half of the input patches are randomly masked during training.
The loss balance parameter $\tau = 0.5$ assigns equal weight to the reconstruction and contrastive objectives, while the perceptual loss weight is $\lambda = 0.1$.
For the perceptual term, we extract \gls{vgg}-16 features from layers $\{\texttt{conv1\_2}, \texttt{conv2\_2}, \texttt{conv3\_3}, \texttt{conv4\_3}\}$, where \texttt{convm\_n} denotes the $n$-th convolutional layer in the $m$-th \gls{vgg} block.
The \gls{ntx} temperature is set to $t = 0.1$.
Contrastive features are extracted from encoder layers $\mathcal{K} = \{3, 6, 9, 12\}$ with corresponding layer weights $\mathbf{w} = (0.1, 0.2, 0.3, 0.4)$.
A Multi-Similarity Miner with margin $\epsilon = 0.1$ is used for hard pair selection.

The model is optimized using AdamW with $\beta_1 = 0.9$, $\beta_2 = 0.95$, a base learning rate of $1 \times 10^{-4}$ and weight decay of $0.1$.
The learning rate follows a cosine annealing schedule with a 10-epoch linear warmup phase.
Gradients are clipped to a maximum norm of 1.0.
Training runs for up to 100 epochs with early stopping triggered after 10 epochs without validation improvement.

A batch contains UV texture maps extracted from frames of $N_{\text{id}} = 8$ distinct identities, with $N_{\text{vid}} = 5$ videos sampled per identity and $N_{\text{frm}} = 5$ frames sampled per video, yielding a total batch size of $B = N_{\text{id}} \times N_{\text{vid}} \times N_{\text{frm}} = 200$ UV texture maps, randomly shuffled within each batch.

During training, augmentation is applied to each frame with Albumentations~\cite{buslaevAlbumentationsFastFlexible2020}, with the following stochastic pipeline:
\begin{itemize}
	\item with probability 0.3, we apply one augmentation among color jitter, Gaussian noise addition
    or Gaussian blur;
	\item with probability 0.5, we apply JPEG compression at quality $q \in [30, 90]$.
\end{itemize}

To promote generalization and to speed up the training, at each epoch only a random subset of available identities is used for training, controlled by a sampling ratio of 0.5.
This means that each epoch trains on 50\% of the total identities, with different random subsets selected across epochs.

\section{Results}
\label{sec:results}

In this section, we first evaluate the best configuration of our method in terms of performance with respect to the reference-set size.
We then compare against the state of the art in terms of generalization, robustness, threshold stability and computation time.
Finally,
we show results related to our proposed interpretability analysis.

\vspace{-10pt}
\subsection{Reference Set Size}
\label{subsec:ref-size}

In the \gls{poi} setting, the reference set is a crucial component of the detection pipeline, as it determines the latent distribution of the identity of the subject of interest.
To assess how the size of the reference set affects performance, we consider a variable number of reference videos $N_{\text{vid}}$ and a variable number of sampled frames per video $N_{\text{frm}}$.

\begin{figure}[t]
\centering
\includegraphics[width=.98\columnwidth]{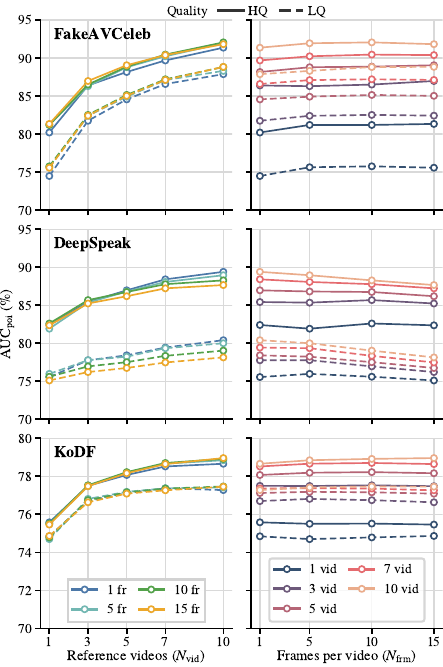}
\caption{Effect of the reference-set size on $\mathrm{AUC}_{\mathrm{poi}}$ across FakeAVCeleb, DeepSpeak and KoDF, under both \gls{hq} and \gls{lq} settings.}
\label{fig:ref-size-grid}
\vspace{-15pt}
\end{figure}

Fig.~\ref{fig:ref-size-grid} compares the \gls{poi}-level metric $\mathrm{AUC}_{\mathrm{poi}}$
under both \gls{hq} and \gls{lq} settings. In the left column, the horizontal axis varies the number of reference videos $N_{\text{vid}}$, while each curve corresponds to a fixed number of sampled frames per video $N_{\text{frm}}$. In the right column, the roles are reversed: the horizontal axis varies $N_{\text{frm}}$, while each curve corresponds to a fixed $N_{\text{vid}}$.
The DF-TIMIT dataset is omitted because it appears to be a comparatively less challenging dataset in our setting, with performance saturated at $100\%$ across all tested configurations.
Consequently, the effect of the reference-set size would be largely uninformative in this analysis.

Focusing on the left column, we observe that the number of reference videos is the factor associated with the clearest performance variation.
Across datasets, increasing the number of videos generally improves performance in both \gls{hq} and \gls{lq} conditions.
On FakeAVCeleb, moving from 1 to 10 reference videos with 1 frame per video yields an $\mathrm{AUC}_{\mathrm{poi}}$ gain of
$11$ points in the \gls{hq} setting
and
more than $13$ points in the \gls{lq} setting.
On DeepSpeak, the same setting yields an $\mathrm{AUC}_{\mathrm{poi}}$ gain of $7$ points in the \gls{hq} setting
and almost $5$
points in the \gls{lq} setting.
KoDF follows the same overall pattern, although with smaller gains.
These results indicate that video-level variability is more valuable than repeatedly sampling many frames from a small number of clips, since additional videos contribute to broader variation in facial appearance and capture conditions.

Considering the right column, the influence of the number of frames per video is weaker and less monotonic.
On FakeAVCeleb, $\mathrm{AUC}_{\mathrm{poi}}$ shows an increasing trend, even though the difference with respect to using 1 frame is very small and does not amount to a real gain.
On DeepSpeak, somewhat unexpectedly, fewer frames are generally preferable, even though the gap between using a single frame and using 15 frames remains marginal.
KoDF is even flatter, with only negligible changes as the number of sampled frames increases.
We believe this behavior is consistent with the use of UV texture maps, which already normalize much of the pose and expression variability, making additional sampling redundant.
This is advantageous because it allows us to sample fewer textures from each reference clip while still building an effective and compact representation of the \gls{poi}.

Overall, results show that the reference set should prioritize diversity across videos.
Using more videos is consistently beneficial in both \gls{hq} and \gls{lq} settings, whereas sampling more frames per video yields only marginal gains and can even be detrimental.

Given these considerations, for the remainder of this article we use 10 reference videos.
Regarding the number of frames, we select 10 frames per video.
Indeed, if not paramount as far as deepfake detection performance is concerned, the choice of using $N_{\text{frm}} = 10$ has been largely dictated by the need to build a sufficiently rich reference support for the interpretability procedure described in Section~\ref{subsec:method_interpretability-mm-mae}.
In particular, our proposed interpretability analysis relies on the estimation of the reference centroid and on the aggregation of multiple reference-to-reference and reference-to-test displacements.
Using 10 frames provides enough within-subject variability to make these quantities more stable, while preserving the compactness of the reference set. We provide more insights on these final considerations in Section~\ref{subsec:interpretability-results}.

\vspace{-10pt}
\subsection{Comparison with State of the Art}
\label{subsec:results_sota}

In this section, we compare our method with the state of the art across several datasets.
Table~\ref{tab:auc_poi} reports the \gls{poi}-level metrics $\mathrm{AUC}_{\mathrm{poi}}$ and $\mathrm{BA}_{\mathrm{poi}}$, while Table~\ref{tab:auc-dataset} reports the dataset-level metrics $\mathrm{AUC}$ and $\mathrm{BA}$.
In both tables, the last column reports the mean across datasets, and the results are organized in two blocks, one for \gls{hq} and one for \gls{lq} evaluation, to highlight the effect of quality degradation.
We also highlight general deepfake detection methods at the top and \gls{poi}-specific methods at the bottom, to better contextualize our method among the baselines.

\begin{table*}[t]
  \centering
  \caption{\gls{poi}-level metrics $(\mathrm{AUC}_{\mathrm{poi}} / \mathrm{BA}_{\mathrm{poi}})$. The last column reports the mean across datasets. Bold indicates the best result among the \gls{poi}-specific methods, while italics indicate the best result among the general deepfake detection methods.}
  \label{tab:auc_poi}
  \begin{tabular}{llccccc}
    \toprule
    & Method & DF-TIMIT & FakeAVCeleb & DeepSpeak & KoDF & Mean \\
    \midrule

    \multirow{6}{*}{\rotatebox[origin=c]{90}{\textit{Higher Quality}}}
    & RealForensics~\cite{haliassosLeveragingRealTalking2022b}
    & \textit{100.00/100.00} & \textit{96.72/95.12} & 85.30/83.96 & 92.07/88.65 & 93.52/91.93 \\

    & LipForensics~\cite{haliassosLipsDontLie2021a}
    & 97.92/98.12 & 96.69/94.36 & \textit{91.69/91.50} & 91.92/89.33 & \textit{94.55/93.33} \\

    & FTCN~\cite{zhengExploringTemporalCoherence2021}
    & \textit{100.00/100.00} & 83.99/84.29 & 76.65/82.17 & \textit{92.21/89.16} & 88.21/88.90 \\

    & Seferbekov~\cite{seferbekovSelimsefDfdc_deepfake_challenge2026}
    & 90.00/92.50 & 96.48/94.60 & 58.42/69.24 & 87.74/86.68 & 83.16/85.76 \\

    \cmidrule(lr){2-7}

    & POI-Forensics~\cite{cozzolinoAudioVisualPersonofInterestDeepFake2023}
    & 89.58/93.33 & 78.59/80.68 & 71.79/75.63 & \textbf{83.48/81.28} & 80.86/82.73 \\

    & \glsentryshort{cupid} (ours)
    & \textbf{100.00/100.00} & \textbf{90.02/89.36} & \textbf{88.12/85.61} & 78.94/78.24 & \textbf{89.27/88.30} \\

    \midrule

    \multirow{6}{*}{\rotatebox[origin=c]{90}{\textit{Lower Quality}}}
    & RealForensics~\cite{haliassosLeveragingRealTalking2022b}
    & 76.98/84.43 & 66.22/70.44 & \textit{60.89/68.10} & 79.60/77.29 & \textit{70.92/75.07} \\

    & LipForensics~\cite{haliassosLipsDontLie2021a}
    & 71.77/79.01 & 68.60/71.76 & 51.08/62.12 & \textit{83.17/81.00} & 68.66/73.47 \\

    & FTCN~\cite{zhengExploringTemporalCoherence2021}
    & \textit{87.29/88.65} & 28.30/55.56 & 59.12/67.20 & 62.17/64.20 & 59.22/68.90 \\

    & Seferbekov~\cite{seferbekovSelimsefDfdc_deepfake_challenge2026}
    & 53.80/70.05 & \textit{81.20/80.84} & 50.40/61.65 & 68.46/70.79 & 63.47/70.83 \\

    \cmidrule(lr){2-7}

    & POI-Forensics~\cite{cozzolinoAudioVisualPersonofInterestDeepFake2023}
    & 89.48/93.23 & 78.33/80.24 & 69.01/72.74 & \textbf{82.66/80.54} & 79.87/81.69 \\

    & \glsentryshort{cupid} (ours)
    & \textbf{100.00/100.00} & \textbf{87.47/86.34} & \textbf{79.19/77.99} & 77.63/77.48 & \textbf{86.07/85.45} \\

    \bottomrule
  \end{tabular}
\end{table*}

\begin{table*}[t]
  \centering
  \caption{Dataset-level metrics $(\mathrm{AUC} / \mathrm{BA})$. The last column reports the mean across datasets. Bold indicates the best result among the \gls{poi}-specific methods, while italics indicate the best result among the general deepfake detection methods.}
  \label{tab:auc-dataset}
  \begin{tabular}{llccccc}
    \toprule
    & Method & DF-TIMIT & FakeAVCeleb & DeepSpeak & KoDF & Mean \\
    \midrule

    \multirow{6}{*}{\rotatebox[origin=c]{90}{\textit{Higher Quality}}}
    & RealForensics~\cite{haliassosLeveragingRealTalking2022b}
    & 99.63/97.71 & \textit{97.21/91.66} & 81.50/72.56 & 87.91/80.04 & 91.56/85.49 \\

    & LipForensics~\cite{haliassosLipsDontLie2021a}
    & 97.08/92.19 & 96.81/90.35 & \textit{89.36/82.94} & 85.03/77.15 & \textit{92.07/85.66} \\

    & FTCN~\cite{zhengExploringTemporalCoherence2021}
    & \textit{100.00/100.00} & 82.51/75.74 & 77.01/69.31 & \textit{88.91/80.91} & 87.11/81.49 \\

    & Seferbekov~\cite{seferbekovSelimsefDfdc_deepfake_challenge2026}
    & 85.46/81.61 & 96.74/90.76 & 54.53/55.45 & 85.54/80.22 & 80.57/77.01 \\

    \cmidrule(lr){2-7}

    & POI-Forensics~\cite{cozzolinoAudioVisualPersonofInterestDeepFake2023}
    & 89.11/87.76 & 77.85/71.40 & 66.69/66.03 & \textbf{84.09/77.70} & 79.44/75.72 \\

    & \glsentryshort{cupid} (ours)
    & \textbf{100.00/100.00} & \textbf{88.63/80.52} & \textbf{84.21/76.78} & 78.94/74.79 & \textbf{87.94/83.02} \\

    \midrule

    \multirow{6}{*}{\rotatebox[origin=c]{90}{\textit{Lower Quality}}}
    & RealForensics~\cite{haliassosLeveragingRealTalking2022b}
    & 73.54/68.87 & 65.74/61.54 & \textit{58.38/56.78} & 75.78/69.10 & \textit{68.36/64.07} \\

    & LipForensics~\cite{haliassosLipsDontLie2021a}
    & 68.88/65.73 & 67.70/63.47 & 51.12/51.79 & \textit{76.73/69.86} & 66.11/62.71 \\

    & FTCN~\cite{zhengExploringTemporalCoherence2021}
    & \textit{83.07/76.56} & 28.19/50.12 & 57.14/56.06 & 59.98/57.02 & 57.09/59.94 \\

    & Seferbekov~\cite{seferbekovSelimsefDfdc_deepfake_challenge2026}
    & 52.91/55.10 & \textit{80.96/73.69} & 48.80/51.25 & 68.23/65.38 & 62.73/61.36 \\

    \cmidrule(lr){2-7}

    & POI-Forensics~\cite{cozzolinoAudioVisualPersonofInterestDeepFake2023}
    & 88.46/87.03 & 77.50/71.21 & 64.46/63.54 & \textbf{83.32/77.12} & 78.43/74.73 \\

    & \glsentryshort{cupid} (ours)
    & \textbf{99.97/99.32} & \textbf{85.41/77.58} & \textbf{73.47/66.77} & 77.94/74.44 & \textbf{84.20/79.53} \\

    \bottomrule
  \end{tabular}
\end{table*}

Compared with the general deepfake detectors, \glsentryshort{cupid} is already competitive in the \gls{hq} setting and becomes the strongest solution under quality degradation.

At the \gls{poi} level, LipForensics attains the highest mean $\mathrm{AUC}_{\mathrm{poi}}$ in the \gls{hq} setting, but \glsentryshort{cupid} remains close, trailing by 5.28 points on the mean while preserving strong results on datasets such as DF-TIMIT and DeepSpeak.

The dataset-level comparison shows the same pattern, with LipForensics ahead by 4.13 points on the \gls{hq} mean.
However, the picture changes in the \gls{lq} setting, where \glsentryshort{cupid} becomes the best overall method in both \gls{poi}-level and dataset-level evaluation.
Its mean $\mathrm{AUC}_{\mathrm{poi}}$ decreases by only 3.2 points from \gls{hq} to \gls{lq}, whereas the general deepfake baselines suffer an average drop of about 24.3 points and fall to 65.6 on average in the \gls{lq} setting.

This advantage is accompanied by a more stable behavior with respect to threshold calibration.
Measured through the gap between $\mathrm{BA}_{\mathrm{poi}}$ and $\mathrm{BA}$, the general deepfake baselines exhibit average gaps of 7.57 points in the \gls{hq} setting and 10.05 points in the \gls{lq} setting, whereas \glsentryshort{cupid} achieves smaller gaps of 5.28 and 5.92 points, respectively.
These results indicate that the proposed method not only retains stronger discrimination under compression, but also preserves a more stable operating behavior when moving from subject-specific calibration to a single threshold shared across the dataset.

Focusing on the comparison with the \gls{poi}-specific baseline, our method is consistently stronger than POI-Forensics on almost all datasets and conditions.
This behavior is visible both in the \gls{poi}-level aggregation of Table~\ref{tab:auc_poi} and in the stricter dataset-level aggregation of Table~\ref{tab:auc-dataset},
which suggests that the gain is not tied to the specific thresholding strategy but rather to a stronger underlying signal.
The same trend also emerges in terms of threshold calibration: the gap between $\mathrm{BA}_{\mathrm{poi}}$ and $\mathrm{BA}$ is 7.01 points in the \gls{hq} setting and 6.96 points in the \gls{lq} setting for POI-Forensics, while \glsentryshort{cupid} reduces it to 5.28 and 5.92 points, respectively.
This interpretation is consistent with the detailed threshold comparison we report in Section~\ref{subsec:results_threshold-stability}.

The only systematic exception is the KoDF dataset, where POI-Forensics remains slightly ahead.
We conjecture this might be due to a distribution mismatch between KoDF and the data used during training.
KoDF is a large-scale dataset focused on Korean subjects, a demographic group that is underrepresented in VoxCeleb2, the dataset used for training.
This shows that our identity-based method might be sensitive to ethnicity-related distribution shifts, which is a known challenge in deepfake detection~\cite{trinhExaminationFairnessAI2021b}.

Overall, these results show that \glsentryshort{cupid} provides the most favorable trade-off across the evaluation settings considered.
It remains competitive in the \gls{hq} regime, achieves the strongest overall behavior under compression, and in most cases improves over the other \gls{poi}-specific baseline.
This is particularly relevant because \gls{lq} videos are often the practical scenario in online media, where compression and resizing are common.

Notably, \glsentryshort{cupid} attains this performance despite its weaker supervision setting, being trained without any deepfake samples and remaining \gls{poi}-agnostic.
This suggests that the identity-contextual reconstruction and similarity cues extracted from UV facial representations provide a strong signal for \gls{poi} deepfake detection, even under cross-dataset evaluation and adverse compression conditions.

\vspace{-10pt}
\subsection{Robustness to resizing}
\label{subsec:robustness}
We continue our analysis by addressing robustness against resizing, another common post-processing operation in online media.
For the sake of brevity, we evaluate results only for the two \gls{poi}-specific methods (i.e., ours and POI-Forensics), since general-purpose detectors are already known to be sensitive to this type of perturbation \cite{verdoliva2020media} and the two \gls{poi} methods have already proven more robust than the general methods in the previous section.

\begin{figure}[t]
	\centering
	\includegraphics[width=\columnwidth]{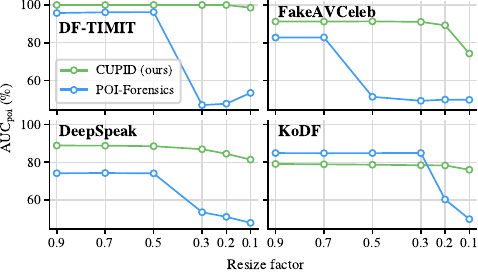}
	\caption{Robustness to spatial downscaling.
	Each plot reports $\mathrm{AUC}_{\mathrm{poi}}$ versus the resize factor.}\label{fig:robustness-resize}
\vspace{-10pt}
\end{figure}

Fig.~\ref{fig:robustness-resize} reports the $\mathrm{AUC}_{\mathrm{poi}}$ under a downscaling sweep, with resize factors ranging from $0.9$ to $0.1$ for all datasets. For brevity, we do not report \acrshort{BApoi} results and dataset-level metrics, since their behavior is consistent with \acrshort{AUCpoi}.\footnote{The corresponding \acrshort{BApoi} and dataset-level results are reported in Figs.~S1 and~S2 of the supplementary material.}

It is worth noting that our method remains stable throughout the full resizing range, degrading smoothly for lower factors.
We consistently outperform POI-Forensics, which exhibits dataset-dependent breakpoints.
KoDF is the only case where POI-Forensics is initially stronger, but its advantage reverses for resizing factors below $0.3$, where \gls{cupid} achieves higher accuracy.
Even though these factors may seem too low and unrealistic, it is important to note that the initial resolution of KoDF is $1920 \times 1080$; therefore, a $\times 0.2$ resize implies a final resolution of $384 \times 216$, which can actually resemble a realistic scenario encountered on social media.

\subsection{POI Threshold Stability}
\label{subsec:results_threshold-stability}

We analyze the threshold stability to understand how much a \gls{poi}-specific detector must be calibrated for the subject or dataset under analysis.
For each subject, we compute the decision threshold that maximizes $\mathrm{BA}_{\mathrm{poi}}$, and we analyze the resulting distribution.
If these thresholds form a compact distribution, a common operating point can be reused with limited subject-specific tuning.
Conversely, broad or heavy-tailed distributions indicate stronger dependence on the \gls{poi}.

\begin{figure}[t]
  \centering
  \includegraphics[width=.98\columnwidth]{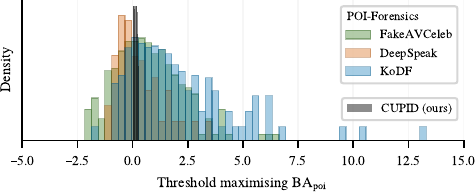}
  \vspace{-15pt}
  \caption{Distribution of per-subject thresholds associated with $\mathrm{BA}_{\mathrm{poi}}$ on the \gls{hq} version of the datasets. For \gls{cupid}, thresholds are pooled across all datasets for visualization purposes.}
  \label{fig:threshold-distributions}
\vspace{-10pt}
\end{figure}

Fig.~\ref{fig:threshold-distributions} reports the threshold distributions for POI-Forensics and for \gls{cupid}.
For convenience, we consider only the \gls{hq} scenario and omit the DF-TIMIT dataset for the same reason given in Section~\ref{subsec:ref-size}.
The figure shows that \gls{cupid} exhibits an extremely compact threshold distribution for all datasets. Note that we aggregated the different datasets for a clearer visualization, since they exhibit
standard deviations in the range $0.014$--$0.042$.
In contrast, POI-Forensics exhibits long-tailed threshold distributions with isolated extreme values, reaching $6.4$ on FakeAVCeleb and $13.4$ on KoDF.
This suggests that \gls{cupid} yields a more stable operating threshold across subjects, making the method easier to calibrate under changing evaluation conditions.

\subsection{Computation Times}
\label{subsec:results_computation-time}
Table~\ref{tab:computation-times} reports statistics of the inference times for 50 subjects of the DeepSpeak dataset ($2092$ videos in total).
We note that \gls{cupid} is $32 \times$ faster than POI-Forensics.

We believe this difference follows from the underlying processing strategy of our method.
For each analyzed video, \gls{cupid} uniformly samples a small, fixed budget of $N_{\text{frm}} = 10$ frames over the whole clip
and runs the encoder once per frame, hence performing exactly $10$ forward passes per video.
This budget does not grow with clip duration.
In fact, since UV texture maps normalize pose and expression, the per-frame identity descriptor is largely redundant across time, so a few frames suffice to capture the identity evidence of the subject.
This justifies such an aggressive subsampling, making inference fast and largely independent of clip length, which also explains the much smaller standard deviation, i.e., more predictable inference times across videos.

By contrast, POI-Forensics adopts a sliding-window strategy, running its network on many short, overlapping temporal segments that span the entire video, so the number of forward passes grows with clip duration.
This makes its inference both more expensive on average and more clip-length dependent, as reflected by its larger standard deviation.

\begin{table}[!t]
  \centering
  \caption{Inference time statistics over 2092 videos (in seconds).}\label{tab:computation-times}
  \begin{tabular}{lcc}
    \toprule
    Method & Mean (s) & Std (s) \\
    \midrule
    \glsentryshort{cupid} & \textbf{0.155} & \textbf{0.043} \\
    POI-Forensics & 4.961 & 5.141 \\
    \bottomrule
  \end{tabular}
\vspace{-10pt}
\end{table}

\subsection{Interpretability Analysis}
\label{subsec:interpretability-results}
For a given test video, \gls{cupid} ultimately returns a single similarity score that drives the real/fake decision.
Beyond this scalar verdict, our detector can also highlight which facial regions deviate from the subject's reference identity, and thus contribute more to the detection score.

Following the procedure described in Section~\ref{subsec:method_interpretability-mm-mae}, we computed the interpretability map $\mathbf{M}^{(\mathcal{T})}$ for all videos in our test datasets.
Here, for brevity's sake, we show interpretability results using two different compact views.

In the first scenario, we randomly select one test identity from each dataset and aggregate the interpretability maps of all its corresponding real and fake videos, obtaining the maps $\mathbf{M}_{\mathrm{id}}^{(\mathrm{real})}$ and $\mathbf{M}_{\mathrm{id}}^{(\mathrm{fake})}$, respectively.
As a second scenario, we compute the average of the resulting interpretability maps over all identities in each dataset, keeping the real and fake sequences separate. This operation yields two maps per dataset, namely $\mathbf{M}^{(\mathrm{real})}$ and $\mathbf{M}^{(\mathrm{fake})}$.

At a dataset level, averaging across subjects allows us to isolate the spatial patterns tied to a forgery family rather than to a specific identity.
Recall that these averaging operations are made possible by working in the UV space: since UV texture maps geometrically align facial content, all residual maps can be meaningfully aggregated.

\begin{figure}[t]
	\centering
	\includegraphics[width=0.90\columnwidth]{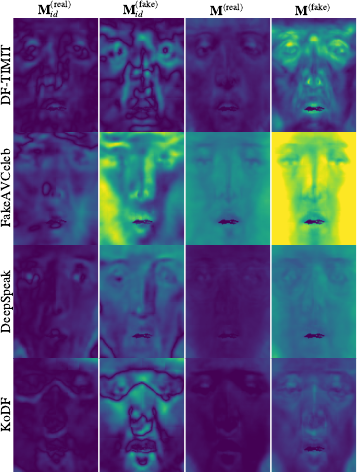}
	\caption{Interpretability residual maps across datasets.
	The four columns report, from left to right, the real interpretability map and the fake interpretability map for a representative identity, and then the average over all subjects of the dataset.
	Brighter regions denote larger deviations from the subject's reference identity.
	All maps share a common dynamic range.}\label{fig:inter_combined}
\vspace{-15pt}
\end{figure}

The resulting maps are reported in Fig.~\ref{fig:inter_combined}.
The four columns show, from left to right, $\mathbf{M}_{\mathrm{id}}^{(\mathrm{real})}$ and $\mathbf{M}_{\mathrm{id}}^{(\mathrm{fake})}$ for a randomly selected identity, followed by $\mathbf{M}^{(\mathrm{real})}$ and $\mathbf{M}^{(\mathrm{fake})}$, the dataset-level averages.
To make residual magnitudes directly comparable, all maps are rendered within a single value range, calibrated between the minimum and maximum residual peaks observed across all datasets.

We start by analyzing the results at the individual-subject level. Consistently across all datasets, the interpretability maps associated with fake videos (second column) exhibit higher activation values than those associated with real videos of the same subject (first column).
These results support the validity of our interpretability analysis: compared to real samples, fake samples produce larger deviations from the subject reference centroid in the latent space.\footnote{Additional video-level examples for further identities of all datasets are reported in Fig.~S3 of the supplementary material.}

$\mathbf{M}_{\mathrm{id}}^{(\mathrm{fake})}$ from DF-TIMIT shows relatively localized peaks around the eyes, nose, and cheeks, consistent with the characteristics of its autoencoder-based face-swap generation process. KoDF also exhibits concentrated responses, with strong activations over the mouth as well, plausibly reflecting its combination of face-swapping and reenactment techniques. FakeAVCeleb displays a more pronounced response near the outer facial boundary, which is consistent with face-swap blending artifacts. Finally, DeepSpeak presents a more diffuse activation pattern. This is consistent with the broader variety of recent generation techniques included in the dataset, particularly avatar-based methods.

The dataset-level averages (third and fourth columns) reinforce the same picture. Here too, for every dataset, the interpretability map associated with real videos consistently exhibits lower values than its fake counterpart.
In line with the single-subject results, DF-TIMIT and KoDF retain sharp, localized peaks around semantically relevant facial regions such as the eyes, nose and mouth. In contrast, FakeAVCeleb and DeepSpeak display more spatially diffuse responses across the face, particularly near its boundaries.
Overall, these qualitative patterns are consistent with the underlying generation pipelines: face-swap-dominated datasets tend to concentrate residuals around identity-bearing landmarks or blending boundaries, whereas more heterogeneous or avatar-style synthesis produces broader deviations.

Several additional observations can be drawn from the overall results. While real maps remain close to the minimum value for almost all datasets, FakeAVCeleb constitutes a notable exception, as its $\mathbf{M}^{(\mathrm{real})}$ attains magnitudes comparable to DeepSpeak's $\mathbf{M}^{(\mathrm{fake})}$.
Moreover, KoDF and DeepSpeak exhibit substantially lower fake residuals than other datasets.
Although we leave a more thorough investigation of FakeAVCeleb's behavior to future work, we believe the different behavior of KoDF and DeepSpeak lies in their recent release: indeed, their lower residuals might reflect the fact that their manipulations are more realistic and thus intrinsically harder to expose.

Even though the substantial differences in residual ranges across datasets might suggest that an interpretability analysis is difficult to apply in practice, these variations are not critical for its practical use. At test time, the actual quantity of interest is the difference between the real and fake residual patterns within each \gls{poi}, rather than the absolute residual magnitude itself.
This contrast remains consistent across subjects in all datasets. It is this relative discrepancy, rather than the absolute residual level, that carries the discriminative signal.

\section{Conclusions}
\label{sec:conclusions}
This paper introduced \glsentryshort{cupid}, a \gls{poi} video deepfake detector that extracts UV texture maps from 3D face reconstructions and feeds them to a \gls{mae} trained only on real videos.
The result is an identity-aware latent space in which a test video can be authenticated by matching its embeddings against those of pristine reference videos of the same \gls{poi}.
Across four datasets and both high- and low-quality settings, it consistently improves over existing methods, it achieves more robust performance and threshold stability, and it is markedly faster than the other \gls{poi}-specific baseline while also being interpretable.

The main limitations suggest two directions for future work.
First, the current video-only representation is mostly identity-centric and does not explicitly model temporal expressivity, so we plan to extend the framework with
objectives that capture subject-specific dynamics to better address LipSync manipulations.
Second, the weaker results on KoDF indicate sensitivity to ethnicity-related distribution shift: training on a more ethnically diverse dataset is necessary to improve fairness and generalization.

\bibliographystyle{IEEEtran}
\bibliography{biblio, biblio2}

\clearpage
\setcounter{section}{0}
\setcounter{figure}{0}
\setcounter{table}{0}
\setcounter{equation}{0}
\renewcommand{\thesection}{S\arabic{section}}
\renewcommand{\thefigure}{S\arabic{figure}}
\renewcommand{\thetable}{S\arabic{table}}
\renewcommand{\theequation}{S\arabic{equation}}

\begin{center}
{\large\bfseries Supplementary Material}
\end{center}
\vspace{4pt}

\noindent
This document provides supplementary material in support of the main paper. The
sections, figures, tables, and equations are intended to complement, not replace, the content of the manuscript.

\section{Best Training Configuration Selection}
\label{subsec:results_ablations}
In this section, we detail the ablation study that motivates the training configuration adopted by \acrshort{cupid}.
Starting from the full \textit{baseline} model, we isolate the impact of its main design choices, namely the balance between the reconstruction and contrastive objectives, the use of canonical UV texture maps instead of face-aligned crops, and the role of data augmentation.
For each variant, we report \gls{auc} and \gls{ba} at both the \gls{poi} and dataset levels, under high- and low-quality settings, and discuss how each choice contributes to detection accuracy and robustness.

Tables~\ref{tab:ablation-auc-mean}--\ref{tab:ablation-auc-dataset} compare five training variants.
The reference model (defined as \textit{baseline}) corresponds to the full training configuration described in Section IV-E in the main paper: UV texture maps as input, mixed objective with $\tau = 0.5$ and $\lambda = 0.1$ (reconstruction, contrastive, and perceptual losses combined), and data augmentation.
Starting from this setting, we define four variants that isolate the contribution of specific design choices:
\begin{itemize}
  \item \textit{no-augmentation}, which is the same as \textit{baseline} but disables data augmentation;
  \item \textit{contrastive-only}, which is the same as the previous variant but sets $\tau = 0$ and $\lambda = 0$, yielding a contrastive-only objective;
  \item \textit{reconstruction-only}, which is the same as \textit{no-augmentation} but sets $\tau = 1$ and $\lambda = 0.1$, thus retaining only the reconstruction branch;
  \item \textit{face-aligned}, which is the same as \textit{no-augmentation} but replaces UV texture maps with face-aligned crops.
\end{itemize}

These variants isolate three design choices: the relative contribution of reconstruction and contrastive learning, the benefit of UV texture maps versus face-aligned crops, and the effect of augmentation on cross-dataset robustness.

\begin{table*}[ht]
  \centering
  \caption{Ablation – \gls{poi}-level metrics $(\mathrm{AUC}_{\mathrm{poi}} / \mathrm{BA}_{\mathrm{poi}})$}
  \label{tab:ablation-auc-mean}
  \begin{tabular}{lccccc}
    \toprule
    Method & DF-TIMIT & FakeAVCeleb & DeepSpeak & KoDF & Mean \\
    \midrule
    \multicolumn{6}{c}{\textit{Higher Quality}} \\
    \midrule
    baseline & \textbf{100.00/100.00} & \textbf{90.02/89.36} & 88.12/85.61 & 78.94/78.24 & 89.27/88.30 \\
    no-augmentation & \textbf{100.00/100.00} & 89.47/88.94 & 88.66/86.16 & 78.18/77.74 & 89.08/88.21 \\
    contrastive-only & \textbf{100.00/100.00} & 89.72/89.08 & \textbf{88.78/86.43} & 78.68/78.03 & \textbf{89.29/88.38} \\
    reconstruction-only & 98.75/99.17 & 67.93/76.18 & 88.15/85.42 & \textbf{79.48/78.30} & 83.58/84.77 \\
    face-aligned & 99.90/99.84 & 86.17/86.43 & 83.90/82.41 & 77.95/77.45 & 86.98/86.53 \\
    \midrule
    \multicolumn{6}{c}{\textit{Lower Quality}} \\
    \midrule
    baseline & \textbf{100.00/100.00} & \textbf{87.47/86.34} & 79.19/77.99 & \textbf{77.63/77.48} & \textbf{86.07/85.45} \\
    no-augmentation & \textbf{100.00/100.00} & 86.31/85.47 & 79.26/78.15 & 76.71/76.81 & 85.57/85.11 \\
    contrastive-only & \textbf{100.00/100.00} & 86.41/85.51 & 79.81/78.50 & 77.24/77.19 & 85.87/85.30 \\
    reconstruction-only & 97.50/97.08 & 66.57/74.99 & \textbf{81.82/79.99} & 76.77/76.70 & 80.67/82.19 \\
    face-aligned & 99.38/99.32 & 84.03/83.96 & 78.82/77.65 & 77.16/76.86 & 84.85/84.45 \\
    \bottomrule
  \end{tabular}
\end{table*}

\begin{table*}[ht]
  \centering
  \caption{Ablation – Dataset-level metrics $(\mathrm{AUC} / \mathrm{BA})$}
  \label{tab:ablation-auc-dataset}
  \begin{tabular}{lccccc}
    \toprule
    Method & DF-TIMIT & FakeAVCeleb & DeepSpeak & KoDF & Mean \\
    \midrule
    \multicolumn{6}{c}{\textit{Higher Quality}} \\
    \midrule
    baseline & \textbf{100.00/100.00} & \textbf{88.63/80.52} & 84.21/76.78 & 78.94/74.79 & \textbf{87.94/83.02} \\
    no-augmentation & \textbf{100.00/100.00} & 88.04/79.85 & 84.22/76.41 & 78.37/74.74 & 87.66/82.75 \\
    contrastive-only & 100.00/99.84 & 87.90/79.71 & 84.47/76.71 & 78.63/74.82 & 87.75/82.77 \\
    reconstruction-only & 98.82/95.94 & 67.27/66.71 & \textbf{84.53/77.27} & \textbf{79.80/74.70} & 82.60/78.65 \\
    face-aligned & 99.71/98.80 & 85.10/77.17 & 79.37/71.97 & 77.91/73.84 & 85.52/80.45 \\
    \midrule
    \multicolumn{6}{c}{\textit{Lower Quality}} \\
    \midrule
    baseline & 99.97/99.32 & \textbf{85.41/77.58} & 73.47/66.77 & \textbf{77.94/74.44} & \textbf{84.20/79.53} \\
    no-augmentation & \textbf{100.00/100.00} & 84.52/76.27 & 73.08/66.17 & 77.20/74.35 & 83.70/79.20 \\
    contrastive-only & 99.97/99.17 & 84.46/76.38 & 73.66/66.79 & 77.53/74.26 & 83.91/79.15 \\
    reconstruction-only & 96.80/90.99 & 66.11/65.62 & \textbf{78.01/71.43} & 77.68/73.76 & 79.65/75.45 \\
    face-aligned & 99.44/97.14 & 82.55/74.60 & 73.49/66.57 & 77.07/73.11 & 83.14/77.85 \\
    \bottomrule
  \end{tabular}
\end{table*}

Focusing only on the two loss variants and the \textit{no-augmentation} configuration, we see from Tables~\ref{tab:ablation-auc-mean} and~\ref{tab:ablation-auc-dataset} that the contrastive signal is the main driver of detection.
\textit{Contrastive-only} obtains the highest mean \gls{auc} in all four summaries and the highest mean \gls{ba} in three of four, while \textit{no-augmentation} remains almost identical.
We believe this behavior can be expected on DF-TIMIT and FakeAVCeleb, whose manipulations alter the subject identity more aggressively.
That is because the contrastive branch is trained to encode identity, and these datasets exhibit a strong identity-mismatch signal due to the aggressive identity alterations.
Conversely, the \textit{reconstruction-only} variant struggles under this assumption, especially on FakeAVCeleb under both compression settings.

The reconstruction branch is nevertheless useful because it captures a complementary, more local signal.
In fact, \textit{reconstruction-only} becomes competitive, and sometimes best, on the DeepSpeak and KoDF datasets.
These datasets contain novel high-quality manipulations that better preserve identity and expose fewer obvious artifacts.
In this regime, identity cues weaken and pixel-level inconsistencies become relatively more informative.

Regarding robustness to quality degradation, although \textit{reconstruction-only} is the weakest solution on average, it drops less than \textit{contrastive-only} when moving from the \gls{hq} to the \gls{lq} setting.
This suggests that reconstruction cues might be more stable under compression and therefore help \textit{no-augmentation} retain the stronger absolute discrimination of the contrastive branch while reducing sensitivity to quality degradation.

Finally, threshold sensitivity can be inferred from the drop between $\mathrm{BA}_{\mathrm{poi}}$ and $\mathrm{BA}$, that is, when moving from one threshold per subject to one threshold per dataset.
By this criterion, \textit{no-augmentation} is the least sensitive overall, with the smallest mean \gls{ba} drop (5.46 and 5.91 points in the \gls{hq} and \gls{lq} settings), suggesting that combining the two signals slightly improves calibration across subjects.

Regarding the input representation, \textit{face-aligned} is consistently worse than the UV-based counterpart under both compression settings and evaluation protocols, indicating that canonical UV texture maps are more informative than face-aligned crops in this \gls{poi} setting, likely because they better normalize pose and spatial correspondence.

Lastly, comparing \textit{baseline} against \textit{no-augmentation} isolates the effect of augmentation: our \textit{baseline}, which enables it, provides the best overall trade-off, yielding the strongest mean results in the \gls{lq} evaluations and at the \gls{hq} dataset level, while remaining essentially tied with \textit{contrastive-only} in the \gls{hq} \gls{poi}-level setting.
This indicates that augmentation improves robustness to quality degradation and distribution shift without sacrificing clean-condition accuracy, which is why it is part of the \textit{baseline} configuration adopted for all the experiments.

\section{Additional Robustness-to-Resizing Results}
\label{sec:supp_robustness}

\begin{figure}[t]
	\centering
	\includegraphics[width=\columnwidth]{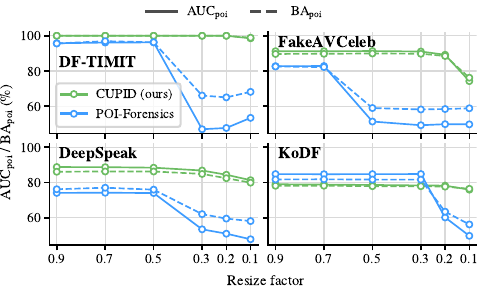}
	\caption{Robustness to spatial downscaling at the \gls{poi} level.
	Each plot reports $\mathrm{AUC}_{\mathrm{poi}}$ (solid) and $\mathrm{BA}_{\mathrm{poi}}$ (dashed) for one dataset over resize factors from $0.9$ to $0.1$.}\label{fig:supp_robustness_poi}
\end{figure}

\begin{figure}[t]
	\centering
	\includegraphics[width=\columnwidth]{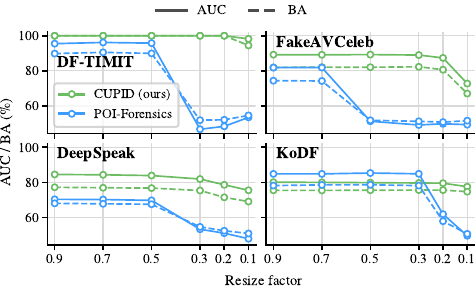}
	\caption{Robustness to spatial downscaling at the dataset level.
	Each plot reports the dataset-level $\mathrm{AUC}$ (solid) and $\mathrm{BA}$ (dashed) for one dataset over resize factors from $0.9$ to $0.1$.}\label{fig:supp_robustness_dataset}
\end{figure}

In the main paper, for brevity, we assess robustness to spatial downscaling only in terms of $\mathrm{AUC}_{\mathrm{poi}}$.
Here we complement that analysis with the metrics omitted there.
Fig.~\ref{fig:supp_robustness_poi} reports the per-subject (\gls{poi}-level) results, showing both $\mathrm{AUC}_{\mathrm{poi}}$ and $\mathrm{BA}_{\mathrm{poi}}$.
Fig.~\ref{fig:supp_robustness_dataset} reports the corresponding dataset-level $\mathrm{AUC}$ and $\mathrm{BA}$.
In both figures, solid and dashed curves denote the AUC and the balanced accuracy, respectively, while colors distinguish the two \gls{poi}-specific methods.
The behavior of these additional metrics closely mirrors that observed for $\mathrm{AUC}_{\mathrm{poi}}$.
\acrshort{cupid} degrades smoothly and remains stable across the full resizing range, whereas POI-Forensics exhibits dataset-dependent breakpoints.
This confirms that the conclusions drawn from $\mathrm{AUC}_{\mathrm{poi}}$ also hold for the balanced accuracy and at the dataset level.

\section{Video-Level Interpretability Examples}
\label{sec:supp_interp_examples}

\begin{figure}[t]
  \centering
  \begin{minipage}{0.49\columnwidth}\centering
    \includegraphics[width=\linewidth]{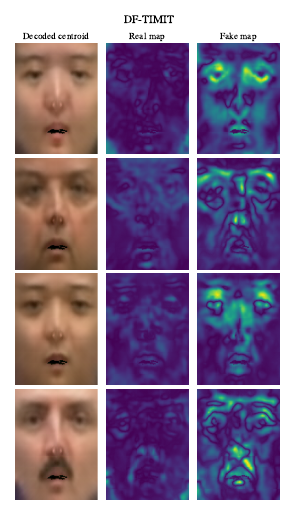}
  \end{minipage}\hfill
  \begin{minipage}{0.49\columnwidth}\centering
    \includegraphics[width=\linewidth]{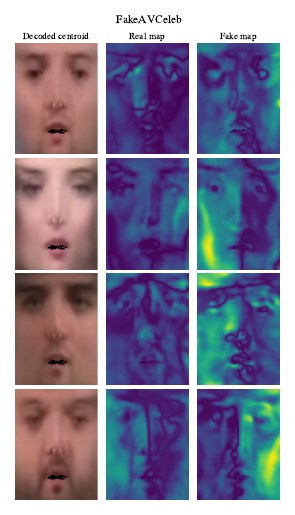}
  \end{minipage}\\[4pt]
  \begin{minipage}{0.49\columnwidth}\centering
    \includegraphics[width=\linewidth]{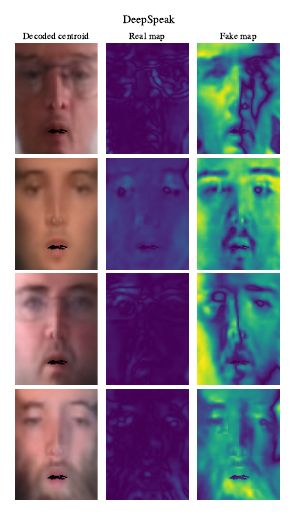}
  \end{minipage}\hfill
  \begin{minipage}{0.49\columnwidth}\centering
    \includegraphics[width=\linewidth]{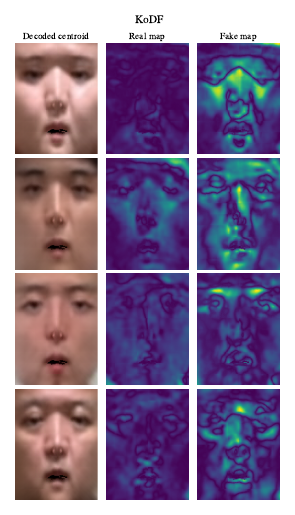}
  \end{minipage}
  \caption{Video-level interpretability examples, with one panel per dataset.
  Each row within a panel corresponds to one identity and shows the decoded reference centroid, the real residual map $\mathbf{M}^{(\mathrm{real})}$, and the fake residual map $\mathbf{M}^{(\mathrm{fake})}$, the last two computed from a single genuine and a single fake video, respectively.
  Each identity uses its own color range, shared between its real and fake map.
  Brighter regions denote larger deviations from the subject's reference identity.}\label{fig:supp_interp_examples}
\end{figure}

The interpretability maps reported in the main paper are aggregated quantities, averaged either over all the videos of a representative identity or over all the identities of a dataset, and are rendered on a single color range shared across datasets so that their absolute magnitudes remain directly comparable.
Here we complement that view with individual, video-level examples.
Fig.~\ref{fig:supp_interp_examples} reports, for each dataset, four identities.
For every identity we show, from left to right, the decoded reference centroid, which represents the subject's average appearance in the UV space, and the residual maps $\mathbf{M}^{(\mathrm{real})}$ and $\mathbf{M}^{(\mathrm{fake})}$ obtained from a single genuine and a single fake video, respectively.
Across all subjects and datasets, the fake map is consistently brighter and more spatially structured than the corresponding real map.
This confirms, at the level of individual videos, the same trend observed on the aggregated maps of the main paper.

\end{document}